\documentclass{article}

\PassOptionsToPackage{numbers, compress}{natbib}
\usepackage[preprint]{neurips_2026}

\usepackage[utf8]{inputenc} 
\usepackage[T1]{fontenc}    
\usepackage{hyperref}       
\usepackage{url}            
\usepackage{booktabs}       
\usepackage{amsfonts}       
\usepackage{nicefrac}       
\usepackage{microtype}      
\usepackage{xcolor}         
\usepackage{enumitem}
\usepackage{graphicx}
\usepackage{pifont}
\usepackage{tabularx}
\usepackage{multirow}
\usepackage{makecell}
\usepackage{amsmath}
\usepackage{tcolorbox}
\usepackage{cleveref}
\usepackage{fontawesome5}
\newcommand{\cmark}{\ding{51}}
\newcommand{\xmark}{\ding{55}}

\title{ESARBench: A Benchmark for Agentic UAV Embodied Search and Rescue}

\author{
    Daoxuan Zhang\hspace{0.3cm}
    Ping Chen\hspace{0.3cm}
    Jianyi Zhou\hspace{0.3cm}
    Shuo Yang$^{\text{\faEnvelope}}$ \\
    \vspace{0.2cm}
    Harbin Institute of Technology, Shenzhen \\
    \vspace{0.2cm}
    \small \texttt{2023311529@stu.hit.edu.cn} \quad
    \small \texttt{shuoyang@hit.edu.cn} \\
}

\begin{document}

\maketitle

\begin{figure*}[h]
    \centering
    \vspace{-10pt}
    \includegraphics[width=1.0\linewidth]{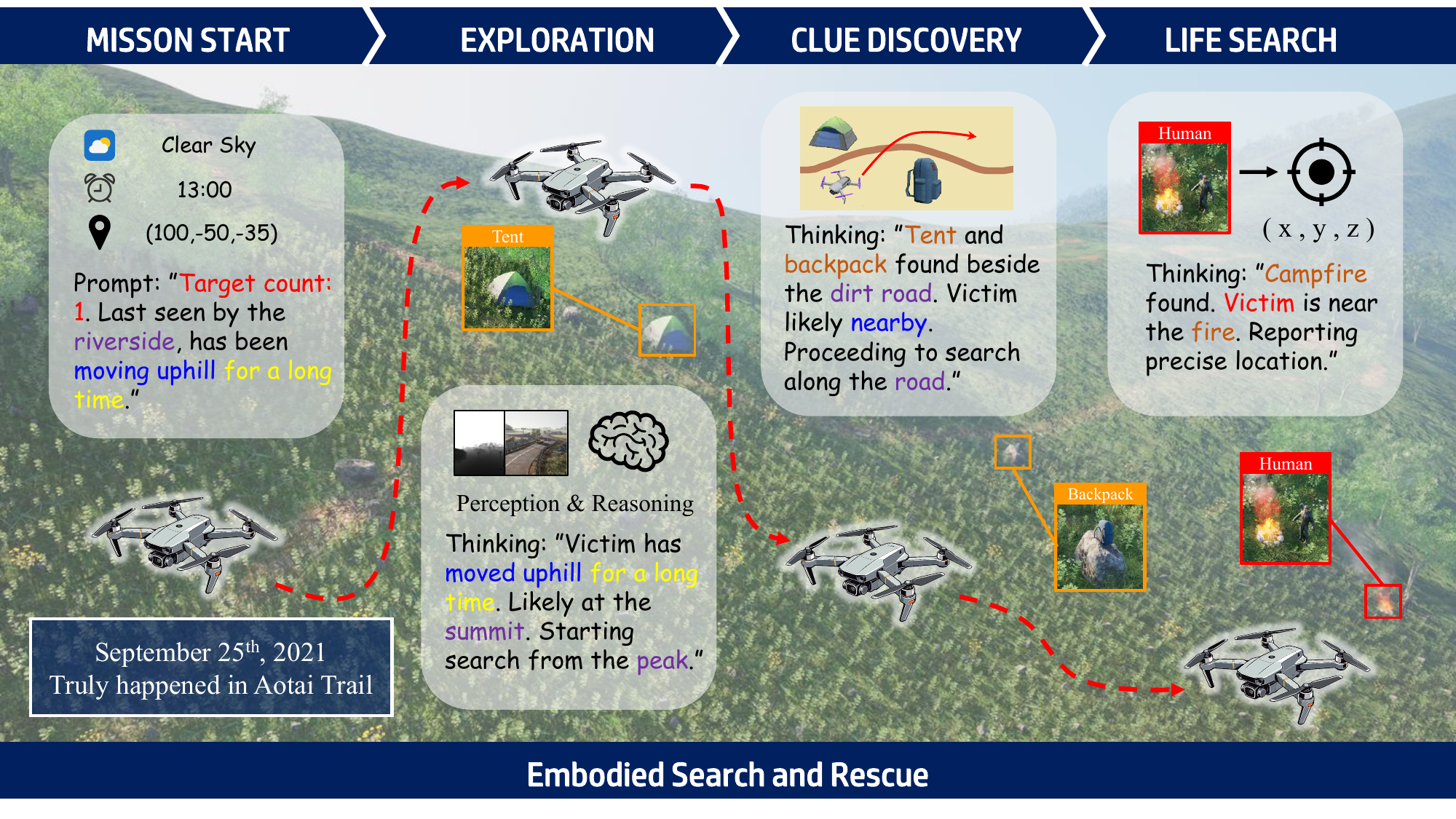}
    \caption{\textbf{Illustration of the Embodied Search and Rescue (ESAR) task workflow.} Modeled after real-world cases, the ESAR mission unfolds across four sequential phases: Mission Start, Exploration, Clue Discovery, and Life Search. The UAV agent is initialized with basic environmental conditions and a textual prompt describing the target's last known trajectory. Throughout the flight, the agent utilizes continuous perception and reasoning to identify vital visual clues, such as a tent and a backpack, dynamically adjusting its search strategy based on these findings. Ultimately, the agent locates the human target and reports precise 3D spatial coordinates.}
    \label{fig:main}
\end{figure*}

\begin{abstract}
  The rapid advancement of Multimodal Large Language Models (MLLMs) has empowered Unmanned Aerial Vehicle (UAV) with exceptional capabilities in spatial reasoning, semantic understanding, and complex decision-making, making them inherently suited for UAV Search and Rescue (SAR). However, existing UAV SAR research is dominated by traditional vision and path-planning methods and lacks a comprehensive and unified benchmark for embodied agents. To bridge this gap, we first propose the novel task of \textbf{Embodied Search and Rescue (ESAR)}, which requires aerial agents to autonomously explore complex environments, identify rescue clues, and reason about victim locations to execute informed decision-making. Additionally, we present \textbf{ESARBench}, the first comprehensive benchmark designed to evaluate MLLM-driven UAV agents in highly realistic SAR scenarios. Leveraging Unreal Engine 5 and AirSim, we construct four high-fidelity, large-scale open environments mapped directly from real-world Geographic Information System (GIS) data to ensure photorealistic landscapes. To rigorously simulate actual rescue operations, our benchmark incorporates dynamic variables including weather conditions, time of day, and stochastic clue placement. Furthermore, we create a dataset of 600 tasks modeled after real-world rescue cases and propose a robust set of evaluation metrics. We evaluate diverse baselines, ranging from traditional heuristics to advanced ground and aerial MLLM-based ObjectNav agents. Experimental results highlight the challenges in ESAR, revealing critical bottlenecks in spatial memory, aerial adaptation, and the trade-off between search efficiency and flight safety. We hope ESARBench serves as a valuable resource to advance research on Embodied Search and Rescue domain. Source code and project page: \url{https://4amgodvzx.github.io/ESAR.github.io}.
\end{abstract}

\section{Introduction}

The integration of Embodied Artificial Intelligence into Unmanned Aerial Vehicles (UAVs) has emerged as a transformative paradigm\citep{Fanglong,embodiedagenteval,vlmsurvey,aeroverse,uavai}, extending the boundaries of robotic autonomy from 2D ground planes\citep{matterport3d,ai2thor,vlnce,navgpt2,voronav,l3mvn} to complex 3D spaces\citep{uavllm,generalpurpose,airvista}. This trend has driven research into various aerial tasks, such as Aerial Vision Language Navigation (VLN)\citep{aerialvln,openuav,citynav} and object goal navigation\citep{uavon,prpsearcher,raven,apex}, which aim to equip drones with the capability to observe, understand, and interact with their environments. Among the many applications of intelligent UAVs\citep{u2udata,u2udata2,rapid,racevla,cognitivedrone}, Search and Rescue (SAR) stands out as a critical domain. Leveraging their flexibility and extensive field of view, UAVs play an indispensable role in disaster response and wilderness rescue\citep{uavvlrr}.

However, two significant gaps hinder the deployment and evaluation of autonomous agents in real-world SAR scenarios.

First, traditional UAV SAR relies on a decoupled stack of classical perception\citep{cnn,drespnet,yolov10} and geometric path planning\citep{realexp,smartagent,cognitive,AUSPEX}. Constrained by a lack of semantic reasoning, these methods depend heavily on narrow, pre-defined operational patterns. This dependency directly results in highly fragmented and task-specific benchmarks, each evaluating custom assumptions rather than generalizable intelligence\citep{drl,coverpath,llmcollabration}. Therefore, they do not provide a unified framework to assess the practical efficacy of UAV agents in real-world SAR tasks.

Moreover, existing embodied UAV researches, such as Aerial VLN\citep{skyvln,navagent,flightgpt,vlfly,openfly}, heavily rely on fine-grained, step-by-step linguistic instructions\citep{openuav,citynavagent,fela,geotext}. These researches lack a high-level, task-centric objective, reducing the agent to a passive instruction follower rather than an active decision-maker. Such setups differ significantly from real-world applications, where instructions are often abstract and goal-oriented. Consequently, current benchmarks fail to comprehensively evaluate a UAV agent's ability to perform long-horizon planning and autonomous exploration under uncertainty\citep{aeroduo,geonav}.

To bridge these gaps, we propose the novel task of \textbf{Embodied Search and Rescue (ESAR)}. As illustrated in \Cref{fig:main}, ESAR requires the agent to demonstrate holistic capabilities, including discovering multi-modal cues, reasoning about environmental semantics, and making autonomous decisions in complex 3D terrains. This task not only meets a vital real-world need but also serves as a benchmark for evaluating the comprehensive capabilities of embodied agents, particularly their potential for task transferability and generalization in open-world settings.

Additionally, to facilitate further research in this domain, we introduce the \textbf{ESARBench}, a high-fidelity simulation platform coupled with a comprehensive evaluation framework for UAV agents. Our primary design objective is to minimize the visual sim-to-real gap, ensuring that agent behaviors learned in simulation are robust and transferable to physical reality. To achieve this, we employ Unreal Engine 5 (UE5) for its high-fidelity rendering capabilities and AirSim\citep{airsim} for accurate flight dynamics and physics simulation.

In this framework, we meticulously constructed four large-scale environments based on real-world Geographic Information System (GIS) data and topographical characteristics. These scenarios replicate four distinct and challenging terrains in China, chosen for their representativeness in real-world SAR incidents: Aotai (High Mountain), Lop Nur (Desert), K2 (Snowy Peaks), and Dapeng (Coast). This diversity ensures that the benchmark tests agents across a broad spectrum of topological and visual conditions. \Cref{tab:comparison} shows the comparison of ESARBench with existing works.

\begin{table*}[t]
\centering
\caption{\textbf{Comparison of ESARBench with Existing Aerial Robotics Benchmarks.} Our benchmark distinguishes itself through high-fidelity simulation, diverse weather conditions, and large-scale real-world GIS data integration.}
\label{tab:comparison}
\resizebox{\textwidth}{!}{%
\begin{tabular}{ccccccc}
\toprule
\textbf{Benchmark} & \textbf{Task} & \textbf{Scenario} & \textbf{Simulation} & \textbf{Map Scale} & \textbf{Weather} & \textbf{Real Data} \\ \midrule
AerialVLN~\citep{aerialvln} & VLN & Urban & UE4 & M & 4 & \xmark \\
TravelUAV~\citep{openuav} & VLN & Urban & UE4 & M & 4 & \xmark \\
UAV-Flow~\citep{uavflow} & VLN/VLA & Urban & UE4 & M & 4 & \xmark \\
UAV-ON~\citep{uavon} & ObjectNav & Urban & UE4 & M & 4 & \xmark \\
LLM-CRF~\citep{llmsar} & SAR & Urban & UE4 & S & 4 & \xmark \\
U2UData~\citep{u2udata} & Perception & Wilderness & UE5 & M & 7 & \cmark \\ \midrule
\textbf{ESAR-Bench} & \textbf{ESAR} & \textbf{Wilderness} & \textbf{UE5} & \textbf{L} & \textbf{13} & \textbf{\cmark} \\ \bottomrule
\end{tabular}%
}
\end{table*}

Going beyond static terrain mapping, we aim to reconstruct authentic rescue narratives. Mission-critical clues such as tents, clothes, and illuminants are distributed across the environments based on the specific ``Event-Snapshot-Task'' generation framework and temporal logic derived from actual rescue cases. Moreover, the benchmark introduces dynamic environmental variables, including shifting weather patterns and time-of-day variations, forcing agents to adapt to changing visibility and lighting conditions.

\begin{figure*}[h]
    \centering
    \includegraphics[width=1.0\linewidth]{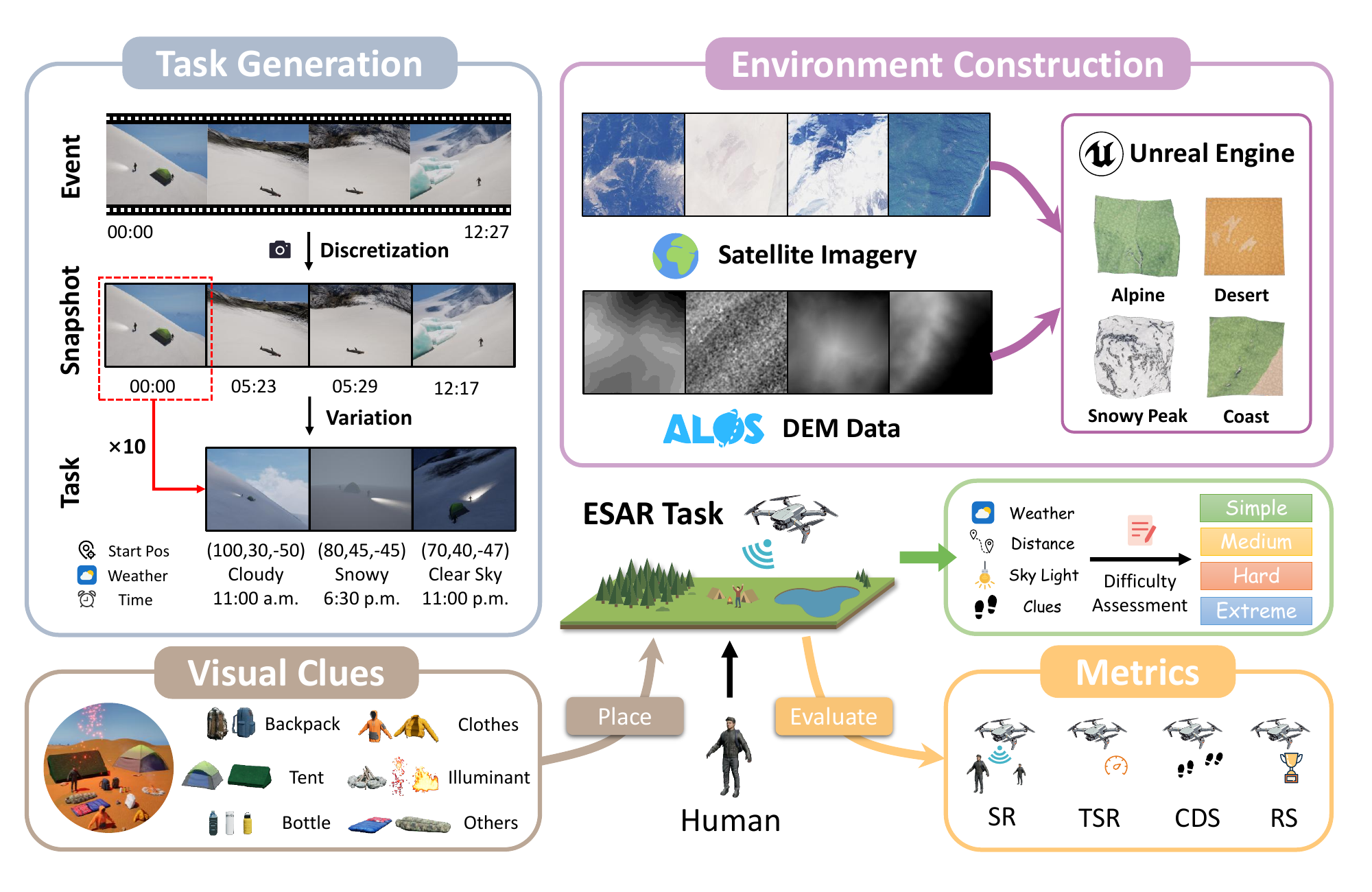}
    \caption{\textbf{Overview of the UAV-ESAR Simulator and Benchmark construction pipeline.} The framework consists of two parallel processes: (1) Environment Construction, which utilizes satellite imagery and DEM data to reconstruct high-fidelity terrains in Unreal Engine 5. (2) Task Generation, which discretizes continuous real-world SAR events into static time snapshots with varying parameters (weather, time of day, starting location). These elements are combined to deploy clues and victims, forming complete ESAR tasks stratified across four difficulty levels. Finally, the performance of embodied UAV agents in these tasks is assessed using SR, TSR, CDS, and RS metrics.}
    \label{fig:method2}
\end{figure*}

Within these dynamic, high-fidelity environments, the UAV agent must process multi-modal sensor inputs to autonomously locate traces of missing persons. To systematically assess their performance in ESAR tasks, we propose a comprehensive suite of evaluation metrics covering Perception Accuracy, Reasoning Capability, and Mission Efficiency. These metrics not only quantify specific SAR performance but also serve to measure the agent's potential for generalizing to other complex embodied tasks.

To establish a baseline for the ESAR task, we evaluate a diverse set of methods, ranging from traditional exploration\citep{frontier,semexp,vlfm} to advanced ground and aerial MLLM-based VLN and ObjectNav agents\citep{navgpt,unigoal,spf,apex} with 3D spatial memory. Our experimental results underscore the significant challenges in ESAR. The findings reveal that direct transfer of ground policies is insufficient; aerial agents require aerial-adapted perception, complex reasoning, and spatial memory, alongside balancing critical trade-offs between search efficiency and flight safety.

In summary, our contributions are:
\begin{itemize}
    \item \textbf{Task Definition:} We are the first to formally propose the concept of Embodied Search and Rescue (ESAR), a novel task designed for practical SAR scenario.
    \item \textbf{Simulation Platform:} We develop the first high-fidelity simulation platform for ESAR agents, featuring photorealistic terrains, dynamic event-driven scenarios, and rich multi-modal sensor interfaces.
    \item \textbf{Benchmark \& Evaluation:} We establish a comprehensive benchmark and evaluation protocol, providing a standardized metric to assess the core cognitive capabilities required for next-generation autonomous UAVs.
\end{itemize}

\section{Related Work}
\subsection{UAVs in Search and Rescue}
Autonomous Search and Rescue (SAR) remains one of the most critical and impactful applications for UAVs. However, existing UAV SAR methodologies heavily relied on traditional perception and geometric path planning\citep{nbv,dual,AUSPEX}. These approaches are often constrained by their dependency on extensive prior knowledge, such as pre-computed environmental maps\citep{smartagent,nbv,dual} or predefined probability models\citep{cognitive,realexp}, which are difficult to acquire in the unpredictable dynamics of real-world emergencies.
Other research streams have focused on isolated computer vision tasks within SAR contexts, such as standalone object detection\citep{yolov10,drespnet} and tracking\citep{cnn}. These vision-only pipelines inherently lack the cognitive functions required for autonomous reasoning, active planning, and high-level decision-making. While Deep Reinforcement Learning (DRL) has been introduced to address some planning limitations\citep{rlmethod,pirlnav,drl,coverpath}, they still struggle with complex semantic reasoning and abstract thinking. Most recently, several studies have integrated Multimodal Large Language Models (MLLMs) into SAR applications\citep{llmcollabration,llmsar,llmselect,SaynFly}, but they still rely on fragmented setups that lack a unified evaluation standard.
In conclusion, the community lacks a high-fidelity simulation platform capable of comprehensively validating the interactive capabilities of these aerial agents. To systematically address these critical voids, we introduce the novel task of Embodied Search and Rescue (ESAR) and provide a premier simulator and benchmark tailored for comprehensive agent evaluation.
\subsection{Embodied Aerial Agents}
To advance the deployment of embodied UAVs into real-world applications\citep{refdrone,selfprompting,avlngrid}, a variety of task paradigms have been proposed to evaluate the capabilities of aerial agents. For instance, Aerial Vision Language Navigation (VLN)\citep{ster,vlaan,longfly,hett,airnav} extends traditional ground-based VLN into 3D environments, assessing an agent's ability to navigate by grounding natural language instructions and visual observations. Specifically, works such as UAV-Flow\citep{uavflow} and SPF\citep{spf} focus on short-horizon navigation guided by concise instructions, while benchmarks like AerialVLN\citep{aerialvln}, TravelUAV\citep{openuav}, CityNav\citep{citynav}, and IndoorUAV\citep{indooruav} emphasize the execution of long-horizon tasks. 
Beyond instruction following, Aerial Object Navigation has been extensively studied, with UAV-ON\citep{uavon}, APEX\citep{apex}, and RAVEN\citep{raven} establishing foundational baselines. Furthermore, datasets like AeroDUO\citep{aeroduo} and U2UData\citep{u2udata} explore the dynamics of multi-UAV collaboration in embodied tasks\citep{uavswarm}. Concurrently, several offline datasets\citep{uavbench,mmuavbench} have been constructed to explicitly evaluate the perception, reasoning, and decision-making capabilities of agents from an aerial perspective.
However, these pioneering works largely focus on isolated sub-tasks lacking direct real-world applicability. Consequently, they fail to adequately assess high-level cognitive capabilities such as complex reasoning and spatial memory, which are essential for practical, end-to-end SAR missions.

\begin{figure*}[t]
    \centering
    \includegraphics[width=1.0\linewidth]{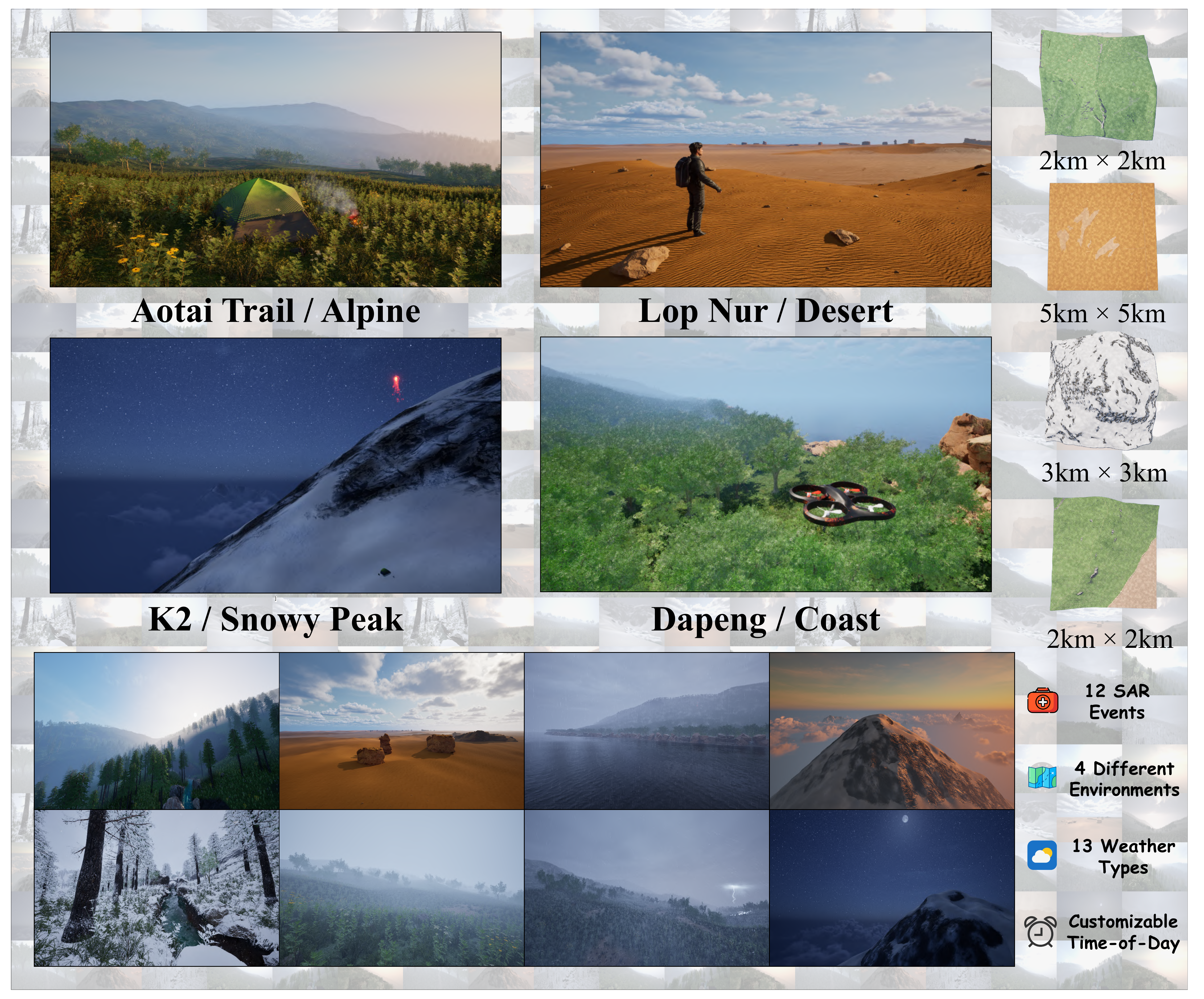}
    \caption{\textbf{Environment Construction and Scenario Variations.} The simulation environments are constructed by integrating real-world GIS data to ensure high terrain fidelity. The figure illustrates four distinct geographic environments with varying physical scales, ranging from $2\text{km} \times 2\text{km}$ to $5\text{km} \times 5\text{km}$. The platform also features dynamic environmental configurations, supporting 13 different weather types and customizable time-of-day settings.}
    \label{fig:method1}
\end{figure*}

\section{UAV-ESAR Simulator and ESARBench}
\label{sec:uav_esar}
\subsection{Task Definition}
In the aerial embodied ESAR task, the UAV agent is required to  navigate complex 3D environments and actively discover mission-critical clues to ultimately locate the victim. At each time step $t$, the agent receives a current visual observation $O_t$, maintains an internal state $S_t$, follows a textual prompt $P$, and leverages historical context $H_t$. Unlike traditional navigation tasks that simply output flight actions, the ESAR agent must also explicitly recognize and output the semantic and spatial information of newly discovered clues, denoted as $M_{t+1}$. Thus, the decision-making process is formulated as a joint output:
\begin{equation}
  A_{t+1}, M_{t+1} = \pi(O_t, S_t, P, H_t).
  \label{eq:taskdef}
\end{equation}
The core objective of the agent is to collect critical clues and locate trapped victims within the shortest possible time. First, a successful victim localization is formally defined when the Euclidean distance between the agent's predicted victim coordinates $C_\text{pred}$ and the actual ground truth coordinates $C_\text{gt}$ is less than or equal to a predefined error threshold $E$:
\begin{equation}
  ||C_\text{pred} - C_\text{gt}||_2 \le E.
  \label{eq:target}
\end{equation}
Second, the agent's environmental reasoning capability is scored based on its clue discovery. Let $\mathcal{M}_\text{gt}$ represent the set of ground-truth clues distributed in the environment, and $\mathcal{M}_\text{pred}$ be the set of clues correctly outputted by the agent during the flight. The task requires the agent to maximize the Clue Recall Rate:
\begin{equation}
    \text{CRR} = C_\text{exact} =\frac{|\mathcal{M}_\text{pred} \cap \mathcal{M}_\text{gt}|}{|\mathcal{M}_\text{gt}|}, 
\end{equation}
which directly serves as a component of the Clue Discovery Score (CDS).

\subsection{UAV-ESAR Simulator}
\subsubsection{Environment Construction.} To authentically replicate real-world SAR scenarios, we selected four geographical hotspots in China renowned for high frequencies of rescue incidents, meticulously mapping to four representative geomorphologies: the Aotai Trail (Alpine), Lop Nur (Desert), K2 (Snowy Peak), and the Dapeng Peninsula (Coastal cliffs). Based on these, as shown in \Cref{fig:method1}, we constructed four large-scale, open-world simulation environments spanning 2km x 2km, 2km x 2km, 3km x 3km, and 5km x 5km. The UAV-ESAR Simulator precisely maps ALOS PALSAR 12m Digital Elevation Model (DEM) data into UE5, generating natural landscapes that strictly adhere to real-world topographical features. By integrating UE5's high-fidelity rendering pipeline with the rigorous flight dynamics of the AirSim-Colosseum plugin, the UAV-ESAR Simulator achieves SOTA performance in embodied rescue simulation.

\subsubsection{Scenario Reproduction.} To reconstruct real-world rescue operations, the simulator deploys victims and 12 types of mission-critical clue models—including tents, backpacks, discarded clothing, campfires, and signal flares—at various strategic locations. Crucially, the spatial distribution and contextual placement of these elements are  deeply rooted in historical, real-world SAR incidents reported in their respective environments.
\subsubsection{Sensors and Environmental Dynamics.} The simulator provides a comprehensive suite of UAV sensors, encompassing an IMU, GPS, LiDAR, alongside multi-view RGB imagery and depth maps. Furthermore, UAV-ESAR supports highly customizable configurations for weather and time of day. Depending on the specific task, the weather can be dynamically altered among 13 distinct types tailored to the specific environmental climate. Importantly, under specific meteorological conditions, the simulated natural landscapes undergo corresponding physical state changes, such as dynamic snow accumulation, water puddles, and dust coverage—thereby comprehensively testing the perceptual robustness of the embodied agents.

\subsection{ESAR Benchmark and Evaluation}
\subsubsection{Task Data Generation.} As demonstrated in \Cref{fig:method2}, The ESARBench dataset is constructed through a structured, three-tier hierarchical generation framework: Event-Snapshot-Task. An Event represents a complete, longitudinal real-world search and rescue incident that unfolds over an extended period. To ensure experimental fairness, stringent control of variables, and algorithmic reproducibility, we discretize each continuous Event into multiple static time snapshots. Within any given snapshot, the spatial distribution of the victims and clues remains stationary, representing a specific developmental stage of the overall rescue timeline. Finally, from each snapshot, we instantiate multiple distinct Tasks by randomly sampling combinations of environmental and initialization parameters, specifically the time of day, weather conditions, and the UAV's starting location. In total, the ESARBench comprises 12 Events, 60 snapshots, and 600 unique tasks. \Cref{fig:realcase} shows four representative event examples.

\begin{figure*}[h]
    \centering
    \includegraphics[width=1.0\linewidth]{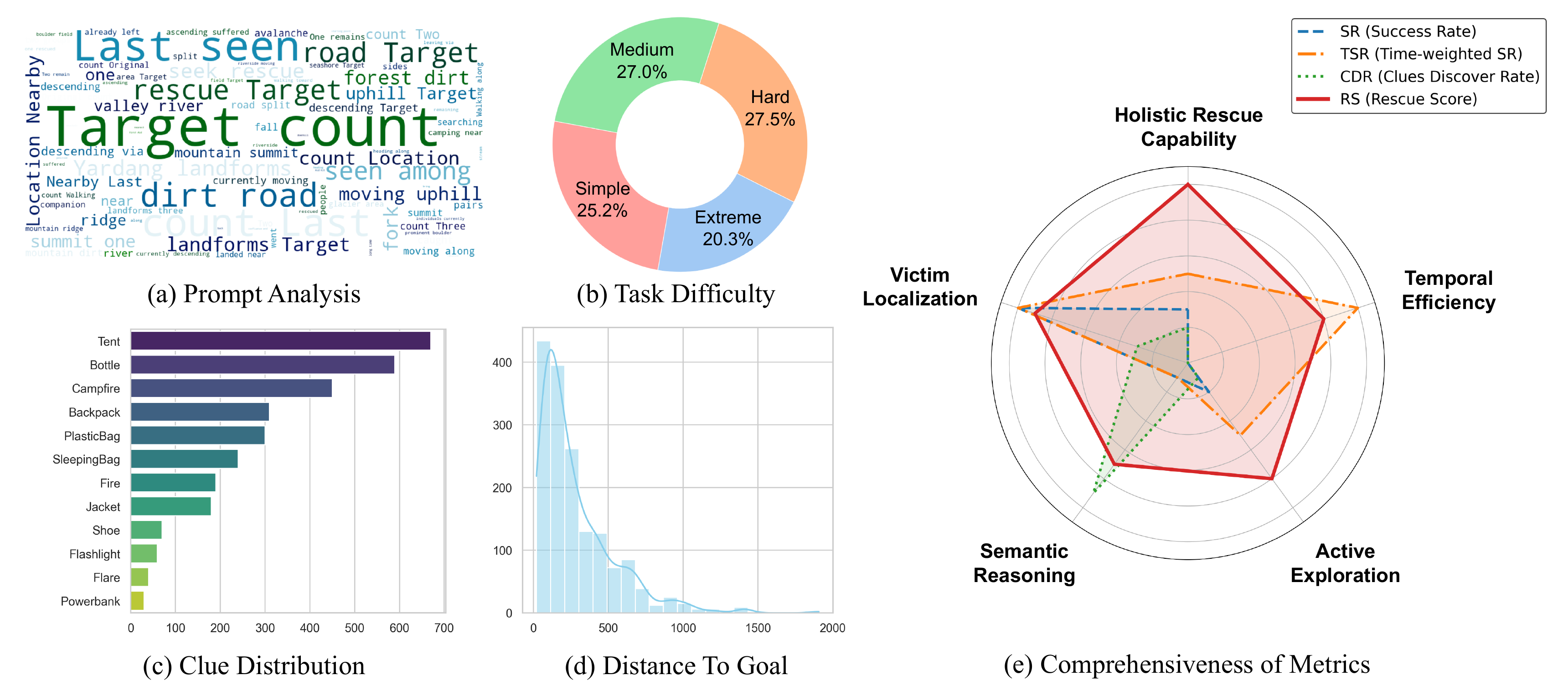}
    \caption{\textbf{Dataset Statistics.} (a) Word cloud analysis of the task prompts. (b) Proportion of tasks across different difficulty levels. (c) Distribution counts of various visual clues. (d) Histogram showing the distribution of initial distances to the goal. (e) Comprehensiveness of our evaluation metrics.}
    \label{fig:data_analysis}
\end{figure*}

\subsubsection{Dataset Stratification.} \Cref{fig:data_analysis} shows the statistics of our task dataset. We stratify the tasks within the ESARBench into four distinct difficulty tiers: Simple, Medium, Hard, and Extreme. This difficulty rating is comprehensively quantified based on a confluence of factors, including weather severity, sky illumination, the average Euclidean distance between the initial starting point and the targets, and the presence of critical clues. Detailed criteria for this difficulty formulation are provided in Appendix A.
\subsubsection{Evaluation Metrics.} To systematically assess the performance of UAV agents, we employ four comprehensive metrics:
\begin{itemize}
    \item \textbf{Success Rate (SR):} The ratio of the number of victims successfully located by the UAV to the total number of victims in the environment. We employ the Hungarian algorithm to compute the optimal bipartite matching between the agent's predicted coordinates and the ground truth locations.
    \begin{equation}
        \text{SR} = \frac{N_\text{found}}{N_\text{total}}.
        \label{eq:tasksr}
    \end{equation}
    \item \textbf{Time-weighted Success Rate (TSR):} A metric that simultaneously evaluates the localization success rate and mission efficiency, where $T$ is the time taken and $T_\text{max}$ is the maximum allowable task time related to the map.
    \begin{equation}
        \text{TSR} = \max \left( 0, SR \times \left( 1 - \frac{T}{T_\text{max}} \right) \right).
        \label{eq:tasktsr}
    \end{equation}
    \item \textbf{Clues Discover Score (CDS):} A comprehensive metric evaluating the UAV's ability to discover mission-critical clues. CDS equally weights pure spatial localization ($C_\text{loc}$) and strict exact matching ($C_\text{exact}$). $C_\text{exact}$ requires both spatial proximity (within threshold $E$) and semantic correctness verified by an LLM evaluator.
    \begin{equation}
        \text{CDS} = 0.5 \cdot \frac{C_\text{loc}}{C_\text{total}} + 0.5 \cdot \frac{C_\text{exact}}{C_\text{total}},
        \label{eq:tasktcds}
    \end{equation}
    \item \textbf{Rescue Score (RS):} A holistic metric designed to evaluate the overall capability of the agent in the ESAR task. It balances the primary objective (finding victims) with safe task completion, semantic exploration (finding clues), and temporal efficiency ($E_t$). The variable $I_{\text{safe}} \in \{0, 1\}$ equals $1$ if the agent safely completes the mission, and $0$ otherwise. We empirically set the weights to $W_{\text{safe}}=0.1$, $W_{\text{base}}=0.3$, $W_{\text{time}}=0.3$, and $W_{\text{clue}}=0.3$. The RS is formulated as:
    \begin{equation}
        \text{RS} = \left[ W_{\text{safe}} \times I_{\text{safe}} \right] + \left[ \text{SR} \times (W_{\text{base}} + W_{\text{time}} \times E_t) \right] + \left[ W_{\text{clue}} \times \text{CDS} \right].
        \label{eq:tasktrs}
    \end{equation}
\end{itemize}

    \begin{table*}[ht]
      \centering
      \setlength{\tabcolsep}{5pt}
      \caption{The performance of different baseline methods in victim searching tasks, including basic methods, ObjectNav methods without MLLMs, and MLLM-based agents adapted from ground or aerial navigation.}
      \begin{tabular}{@{}c*{10}{c}@{}}
      \toprule
      \textbf{Method} & \multicolumn{2}{c}{\textbf{Simple}} & \multicolumn{2}{c}{\textbf{Medium}} & \multicolumn{2}{c}{\textbf{Hard}} & \multicolumn{2}{c}{\textbf{Extreme}} & \multicolumn{2}{c}{\textbf{Overall}}\\
      \cmidrule(lr){2-3} \cmidrule(lr){4-5} \cmidrule(lr){6-7} \cmidrule(lr){8-9} \cmidrule(lr){10-11}
      & \textbf{SR$\uparrow$} & \textbf{TSR$\uparrow$} & \textbf{SR$\uparrow$} & \textbf{TSR$\uparrow$} 
      & \textbf{SR$\uparrow$} & \textbf{TSR$\uparrow$} & \textbf{SR$\uparrow$} & \textbf{TSR$\uparrow$}
      & \textbf{SR$\uparrow$} & \textbf{TSR$\uparrow$}  \\
      \midrule
      Random & 4.68 & 4.44 & 4.24 & 3.87 & 1.04 & 0.98 & 0.00 & 0.00 & 2.65 & 2.47 \\
      FBE\citep{frontier} & 13.33 & 5.64 & 5.83 & 0.86 & 10.52 & 1.54 & 2.31 & 0.00 & 8.19 & 2.05 \\
      Pure-MLLM & 13.21 & 6.57 & 0.00 & 0.00 & 1.04 & 0.83 & 0.00 & 0.00 & 3.45 & 1.80 \\
      SemExp\citep{semexp} & 3.23 & 0.00 & 9.70 & 0.55 & 8.33 & 3.01 & 5.56 & 1.27 & 6.83 & 1.21 \\
      VLFM\citep{vlfm} & 12.10 & 3.98 & 10.51 & 1.69 & 6.35 & 2.53 & 6.96 & 5.12 & 9.12 & 3.17 \\
      NavGPT\citep{navgpt} & 3.23 & 1.87 & 4.65 & 1.78 & 6.35 & 1.45 & 10.56 & 0.00 & 5.92 & 1.36 \\
      UniGoal\citep{unigoal} & 8.06 & 1.75 & 1.82 & 0.00 & 8.96 & 1.02 & 7.50 & 0.10 & 6.47 & 0.74 \\
      SPF\citep{spf} & 17.33 & 1.21 & 4.85 & 0.96 & 7.50 & 0.93 & 5.49 & 0.59 & 8.84 & 0.94 \\
      APEX\citep{apex} & 27.83 & 1.43 & 10.91 & 0.99 & 6.20 & 0.46 & 10.83 & 0.54 & 13.89 & 0.87 \\
      \bottomrule
      \end{tabular}
      \label{tab:tabs}
      \vspace{-10pt}
    \end{table*}

\section{Experiment}
\label{sec:experiment}
\subsection{Baselines}
To establish baselines for the novel ESAR task, we adapt representative methods from several established embodied intelligence settings. These baselines cover several types: basic exploration and direct MLLM control, ObjectNav methods without MLLMs, and MLLM-based agents from ground VLN, ground ObjectNav, aerial VLN, and aerial ObjectNav. For fair comparison, all baselines are connected to the same AirSim interface and use the same four-camera YOLO-World\citep{yoloworld} RGB-D module for clue and victim reporting. For the MLLM components, we use Qwen3.5-Plus \citep{qwen3vl} as our base model.
\begin{itemize}[itemsep=1.0ex]
    \item \textbf{Random (Basic):} A lower-bound policy that uniformly samples discrete UAV actions without mapping or task reasoning.
    \item \textbf{FBE\citep{frontier} (Basic):} A classical frontier-based exploration method using a BEV occupancy map and FMM local planning.
    \item \textbf{Pure-MLLM (Basic):} A direct MLLM-control baseline that maps the front-view observation and task prompt to discrete UAV actions.
    \item \textbf{SemExp\citep{semexp} (Ground ObjectNav, non-MLLM):} A semantic-exploration baseline using a BEV semantic map, frontier selection, and FMM planning. We adapt it to ESAR by replacing the original learned global policy with a zero-shot frontier heuristic.
    \item \textbf{VLFM\citep{vlfm} (Ground ObjectNav, non-MLLM):} A vision-language frontier baseline that ranks frontiers with a value map. It uses image-text matching to estimate which frontier is most relevant to the task prompt.
    \item \textbf{NavGPT\citep{navgpt} (Ground VLN, MLLM):} A ground VLN-style MLLM agent that selects safe actions from multi-view captions, history, and state. It reasons over textual scene descriptions rather than metric maps, serving as a transfer baseline for indoor instruction-following agents in aerial search.
    \item \textbf{UniGoal\citep{unigoal} (Ground ObjectNav, MLLM):} A ground ObjectNav method that uses goal and scene graph matching to guide exploration. It identifies target objects and spatial hints from the SAR prompt, then prioritizes searching in areas that match the current scene graph.
    \item \textbf{SPF\citep{spf} (Aerial VLN, MLLM):} An aerial VLN-style agent that uses VLM point prediction and depth back-projection for point-and-fly control. Unlike ground VLN baselines, it directly predicts an image-space flight direction and converts it into UAV motion commands.
    \item \textbf{APEX\citep{apex} (Aerial ObjectNav, MLLM):} An aerial ObjectNav agent using VLM-guided 3D voxel maps and reward-based discrete action selection. It explicitly models attraction, exploration, and obstacles in 3D space, representing a UAV-specific ObjectNav strategy with spatial memory.
\end{itemize}

\subsection{Results}
\subsubsection{Overall Performance and Aerial Adaptation.} \Cref{tab:tabs,tab:tabc} report the performance of all baselines on victim search, clue discovery, and the comprehensive rescue score. APEX achieves the strongest overall results, ranking first on SR, CDS, and RS with overall scores of 13.89, 4.14, and 13.45, respectively. SPF obtains the second-best RS of 13.12 and remains competitive across both victim search and clue discovery. The advantage of SPF and APEX over the ground MLLM baselines, including NavGPT and UniGoal, indicates that aerial adaptation is crucial for ESAR. UAV agents must handle large-scale outdoor viewpoints, 3D motion, and search-oriented exploration rather than only transferring ground navigation policies.

    \begin{table*}[h]
      \vspace{-5pt}
      \centering
      \setlength{\tabcolsep}{5pt}
      \caption{Baseline experiment results in clue discovery tasks and comprehensive scores (incorporating SR, TSR, and CDS).}
      \begin{tabular}{@{}c*{10}{c}@{}}
      \toprule
      \textbf{Method} & \multicolumn{2}{c}{\textbf{Simple}} & \multicolumn{2}{c}{\textbf{Medium}} & \multicolumn{2}{c}{\textbf{Hard}} & \multicolumn{2}{c}{\textbf{Extreme}} & \multicolumn{2}{c}{\textbf{Overall}}\\
      \cmidrule(lr){2-3} \cmidrule(lr){4-5} \cmidrule(lr){6-7} \cmidrule(lr){8-9} \cmidrule(lr){10-11}
      & \textbf{CDS$\uparrow$} & \textbf{RS$\uparrow$} & \textbf{CDS$\uparrow$} & \textbf{RS$\uparrow$} 
      & \textbf{CDS$\uparrow$} & \textbf{RS$\uparrow$} & \textbf{CDS$\uparrow$} & \textbf{RS$\uparrow$}
      & \textbf{CDS$\uparrow$} & \textbf{RS$\uparrow$}  \\
      \midrule
      Random & 1.25 & 9.55 & 3.49 & 13.08 & 1.04 & 7.48 & 0.0 & 8.75 & 1.51 & 9.81 \\
      FBE & 3.65 & 13.42 & 5.71 & 10.43 & 3.61 & 9.39 & 0.27 & 6.16 & 3.40 & 9.97 \\
      Pure-MLLM & 5.21 & 11.78 & 2.28 & 8.20 & 1.82 & 6.11 & 0.00 & 7.08 & 2.39 & 8.26 \\
      SemExp & 2.14 & 7.72 & 5.50 & 11.85 & 1.80 & 8.32 & 0.00 & 7.88 & 2.47 & 9.05 \\
      VLFM & 4.01 & 12.44 & 4.24 & 12.09 & 1.36 & 7.13 & 1.96 & 10.30 & 2.92 & 10.50 \\
      NavGPT & 4.14 & 10.47 & 2.93 & 11.21 & 1.69 & 9.72 & 4.86 & 12.54 & 3.30 & 10.89 \\
      UniGoal & 4.71 & 10.76 & 1.83 & 8.92 & 3.68 & 9.10 & 1.13 & 8.03 & 2.94 & 9.27 \\
      SPF & 5.89 & 16.94 & 2.22 & 12.35 & 2.94 & 11.54 & 3.08 & 11.49 & 3.53 & 13.12 \\
      APEX & 7.35 & 18.65 & 4.06 & 13.46 & 1.36 & 9.59 & 3.96 & 12.10 & 4.14 & 13.45 \\
      \bottomrule
      \end{tabular}
      \label{tab:tabc}
      \vspace{-10pt}
    \end{table*}

\subsubsection{Semantic Reasoning in Clue Discovery.} MLLM-based methods show a clear advantage in CDS, with the four adapted MLLM baselines averaging 3.48 overall CDS, compared with 2.70 for the non-MLLM ObjectNav baselines. This trend suggests that semantic understanding and reasoning are important for ESAR, where clues are not merely visual objects but task-relevant evidence that must be interpreted under the rescue prompt. The results suggest that MLLM reasoning is only effective when built into embodied search structures: Pure MLLM models struggle, but SPF and APEX perform better because they add UAV-specific actions and spatial mapping to the model's guidance.
\subsubsection{Efficiency Trade-off and Benchmark Difficulty.} Despite their strong RS, SPF and APEX have low TSR scores of 0.94 and 0.87, respectively. This reveals two limitations. First, these agents lack an active mechanism for judging when a multi-target ESAR mission has been sufficiently completed. Second, they struggle to improve efficiency while preserving broad search ability. More generally, all baselines remain far from solving the ESAR Task. Even the best SR and CDS are only 13.89 and 4.14, and the performance gap between the strongest methods and basic baselines is limited. This indicates that ESARBench demands a high level of integration between perception, semantic reasoning, and decision-making.

\section{Conclusion}
\label{sec:conclusion}
In this paper, we introduce the novel paradigm of Embodied Search and Rescue (ESAR), bridging the gap between embodied intelligence and critical real-world aerial missions. To support this, we develop the first high-fidelity simulation platform designed for UAV ESAR agents, and a comprehensive benchmark to quantitatively evaluate their performance.
Our extensive baseline experiments underscore the profound difficulty of the ESAR task. The results demonstrate that autonomous rescue requires an agent's abilities of active perception, semantic reasoning, long-horizon planning, spatial memory, and robust decision-making within complex 3D environments. In the future, we will continuously upgrade the ESARBench, including integrate additional sensor modalities and introduce a broader spectrum of complex task configurations. Furthermore, we will focus on designing more advanced embodied architectures to drive the evolution of fully autonomous aerial rescue systems.

\newpage
\bibliographystyle{abbrvnat}
\bibliography{main.bib}

@String(CVPR  = {IEEE Conf. Comput. Vis. Pattern Recog.})

@String(ICCV  = {Int. Conf. Comput. Vis.})

@String(ECCV  = {Eur. Conf. Comput. Vis.})

@String(ICML  = {Int. Conf. Mach. Learn.})

@String(ICLR  = {Int. Conf. Learn. Represent.})

@String(AAAI  = {AAAI})

@String(CVPR  = {CVPR})

@String(ICCV  = {ICCV})

@String(ECCV  = {ECCV})

@String(ICML  = {ICML})

@String(ICLR  = {ICLR})

@article{embodiedagenteval,
  author       = {Peter Anderson and
                  Angel X. Chang and
                  Devendra Singh Chaplot and
                  Alexey Dosovitskiy and
                  Saurabh Gupta and
                  Vladlen Koltun and
                  Jana Kosecka and
                  Jitendra Malik and
                  Roozbeh Mottaghi and
                  Manolis Savva and
                  Amir R. Zamir},
  title        = {On Evaluation of Embodied Navigation Agents},
  journal      = {CoRR},
  volume       = {abs/1807.06757},
  year         = {2018},
  eprinttype    = {arXiv},
  eprint       = {1807.06757},
  bibsource    = {dblp computer science bibliography, https://dblp.org}
}

@inproceedings{matterport3d,
  author       = {Angel X. Chang and
                  Angela Dai and
                  Thomas A. Funkhouser and
                  Maciej Halber and
                  Matthias Nie{\ss}ner and
                  Manolis Savva and
                  Shuran Song and
                  Andy Zeng and
                  Yinda Zhang},
  title        = {Matterport3D: Learning from {RGB-D} Data in Indoor Environments},
  booktitle    = {2017 International Conference on 3D Vision, 3DV 2017, Qingdao, China,
                  October 10-12, 2017},
  pages        = {667--676},
  publisher    = {{IEEE} Computer Society},
  year         = {2017},
  doi          = {10.1109/3DV.2017.00081},
  bibsource    = {dblp computer science bibliography, https://dblp.org}
}

@article{ai2thor,
  author       = {Eric Kolve and
                  Roozbeh Mottaghi and
                  Daniel Gordon and
                  Yuke Zhu and
                  Abhinav Gupta and
                  Ali Farhadi},
  title        = {{AI2-THOR:} An Interactive 3D Environment for Visual {AI}},
  journal      = {CoRR},
  volume       = {abs/1712.05474},
  year         = {2017},
  eprinttype    = {arXiv},
  eprint       = {1712.05474},
  bibsource    = {dblp computer science bibliography, https://dblp.org}
}

@inproceedings{navgpt2,
  author       = {Gengze Zhou and
                  Yicong Hong and
                  Zun Wang and
                  Xin Eric Wang and
                  Qi Wu},
  editor       = {Ales Leonardis and
                  Elisa Ricci and
                  Stefan Roth and
                  Olga Russakovsky and
                  Torsten Sattler and
                  G{\"{u}}l Varol},
  title        = {NavGPT-2: Unleashing Navigational Reasoning Capability for Large Vision-Language
                  Models},
  booktitle    = {Computer Vision - {ECCV} 2024 - 18th European Conference, Milan, Italy,
                  September 29-October 4, 2024, Proceedings, Part {VII}},
  series       = {Lecture Notes in Computer Science},
  volume       = {15065},
  pages        = {260--278},
  publisher    = {Springer},
  year         = {2024},
  doi          = {10.1007/978-3-031-72667-5\_15},
  bibsource    = {dblp computer science bibliography, https://dblp.org}
}

@inproceedings{voronav,
  author       = {Pengying Wu and
                  Yao Mu and
                  Bingxian Wu and
                  Yi Hou and
                  Ji Ma and
                  Shanghang Zhang and
                  Chang Liu},
  title        = {VoroNav: Voronoi-based Zero-shot Object Navigation with Large Language
                  Model},
  booktitle    = {Forty-first International Conference on Machine Learning, {ICML} 2024,
                  Vienna, Austria, July 21-27, 2024},
  publisher    = {OpenReview.net},
  year         = {2024},
  bibsource    = {dblp computer science bibliography, https://dblp.org}
}

@article{uavon,
  author       = {Jianqiang Xiao and
                  Yuexuan Sun and
                  Yixin Shao and
                  Boxi Gan and
                  Rongqiang Liu and
                  Yanjing Wu and
                  Weili Guan and
                  Xiang Deng},
  title        = {{UAV-ON:} {A} Benchmark for Open-World Object Goal Navigation with
                  Aerial Agents},
  journal      = {CoRR},
  volume       = {abs/2508.00288},
  year         = {2025},
  doi          = {10.48550/ARXIV.2508.00288},
  eprinttype    = {arXiv},
  eprint       = {2508.00288},
  bibsource    = {dblp computer science bibliography, https://dblp.org}
}

@article{apex,
  author       = {Daoxuan Zhang and
                  Ping Chen and
                  Xiaobo Xia and
                  Xiu Su and
                  Ruichen Zhen and
                  Jianqiang Xiao and
                  Shuo Yang},
  title        = {{APEX:} {A} Decoupled Memory-based Explorer for Asynchronous Aerial
                  Object Goal Navigation},
  journal      = {CoRR},
  volume       = {abs/2602.00551},
  year         = {2026},
  doi          = {10.48550/ARXIV.2602.00551},
  eprinttype   = {arXiv},
  eprint       = {2602.00551},
  bibsource    = {dblp computer science bibliography, https://dblp.org}
}

@inproceedings{u2udata,
  author       = {Tongtong Feng and
                  Xin Wang and
                  Feilin Han and
                  Leping Zhang and
                  Wenwu Zhu},
  editor       = {Jianfei Cai and
                  Mohan S. Kankanhalli and
                  Balakrishnan Prabhakaran and
                  Susanne Boll and
                  Ramanathan Subramanian and
                  Liang Zheng and
                  Vivek K. Singh and
                  Pablo C{\'{e}}sar and
                  Lexing Xie and
                  Dong Xu},
  title        = {U2UData: {A} Large-scale Cooperative Perception Dataset for Swarm
                  UAVs Autonomous Flight},
  booktitle    = {Proceedings of the 32nd {ACM} International Conference on Multimedia,
                  {MM} 2024, Melbourne, VIC, Australia, 28 October 2024 - 1 November
                  2024},
  pages        = {7600--7608},
  publisher    = {{ACM}},
  year         = {2024},
  doi          = {10.1145/3664647.3681151},
  bibsource    = {dblp computer science bibliography, https://dblp.org}
}

@article{u2udata2,
  author       = {Tongtong Feng and
                  Xin Wang and
                  Feilin Han and
                  Leping Zhang and
                  Wenwu Zhu},
  title        = {U2UData-2: {A} Scalable Swarm UAVs Autonomous Flight Dataset for Long-horizon
                  Tasks},
  journal      = {CoRR},
  volume       = {abs/2509.00055},
  year         = {2025},
  doi          = {10.48550/ARXIV.2509.00055},
  eprinttype    = {arXiv},
  eprint       = {2509.00055},
  bibsource    = {dblp computer science bibliography, https://dblp.org}
}

@article{prpsearcher,
  author       = {Yatai Ji and
                  Zhengqiu Zhu and
                  Yong Zhao and
                  Beidan Liu and
                  Chen Gao and
                  Yihao Zhao and
                  Sihang Qiu and
                  Yue Hu and
                  Quanjun Yin and
                  Yong Li},
  title        = {Towards Autonomous {UAV} Visual Object Search in City Space: Benchmark
                  and Agentic Methodology},
  journal      = {CoRR},
  volume       = {abs/2505.08765},
  year         = {2025},
  doi          = {10.48550/ARXIV.2505.08765},
  eprinttype    = {arXiv},
  eprint       = {2505.08765},
  bibsource    = {dblp computer science bibliography, https://dblp.org}
}

@article{uavswarm,
  author       = {Yukai Hou and
                  Jin Zhao and
                  Rongqing Zhang and
                  Xiang Cheng and
                  Liuqing Yang},
  title        = {{UAV} Swarm Cooperative Target Search: {A} Multi-Agent Reinforcement
                  Learning Approach},
  journal      = {{IEEE} Trans. Intell. Veh.},
  volume       = {9},
  number       = {1},
  pages        = {568--578},
  year         = {2024},
  doi          = {10.1109/TIV.2023.3316196},
  bibsource    = {dblp computer science bibliography, https://dblp.org}
}

@inproceedings{pirlnav,
  author       = {Ram Ramrakhya and
                  Dhruv Batra and
                  Erik Wijmans and
                  Abhishek Das},
  title        = {PIRLNav: Pretraining with Imitation and {RL} Finetuning for {OBJECTNAV}},
  booktitle    = {{IEEE/CVF} Conference on Computer Vision and Pattern Recognition,
                  {CVPR} 2023, Vancouver, BC, Canada, June 17-24, 2023},
  pages        = {17896--17906},
  publisher    = {{IEEE}},
  year         = {2023},
  doi          = {10.1109/CVPR52729.2023.01716},
  bibsource    = {dblp computer science bibliography, https://dblp.org}
}

@inproceedings{aerialvln,
  author       = {Shubo Liu and
                  Hongsheng Zhang and
                  Yuankai Qi and
                  Peng Wang and
                  Yanning Zhang and
                  Qi Wu},
  title        = {AerialVLN: Vision-and-Language Navigation for UAVs},
  booktitle    = {{IEEE/CVF} International Conference on Computer Vision, {ICCV} 2023,
                  Paris, France, October 1-6, 2023},
  pages        = {15338--15348},
  publisher    = {{IEEE}},
  year         = {2023},
  doi          = {10.1109/ICCV51070.2023.01411},
  bibsource    = {dblp computer science bibliography, https://dblp.org}
}

@inproceedings{openuav,
  author       = {Xiangyu Wang and
                  Donglin Yang and
                  Ziqin Wang and
                  Hohin Kwan and
                  Jinyu Chen and
                  Wenjun Wu and
                  Hongsheng Li and
                  Yue Liao and
                  Si Liu},
  title        = {Towards Realistic {UAV} Vision-Language Navigation: Platform, Benchmark,
                  and Methodology},
  booktitle    = {The Thirteenth International Conference on Learning Representations,
                  {ICLR} 2025, Singapore, April 24-28, 2025},
  publisher    = {OpenReview.net},
  year         = {2025},
  bibsource    = {dblp computer science bibliography, https://dblp.org}
}

@article{skyvln,
  author       = {Tianshun Li and
                  Tianyi Huai and
                  Zhen Li and
                  Yichun Gao and
                  Haoang Li and
                  Xinhu Zheng},
  title        = {SkyVLN: Vision-and-Language Navigation and {NMPC} Control for UAVs
                  in Urban Environments},
  journal      = {CoRR},
  volume       = {abs/2507.06564},
  year         = {2025},
  doi          = {10.48550/ARXIV.2507.06564},
  eprinttype    = {arXiv},
  eprint       = {2507.06564},
  bibsource    = {dblp computer science bibliography, https://dblp.org}
}

@article{navagent,
  author       = {Youzhi Liu and
                  Fanglong Yao and
                  Yuanchang Yue and
                  Guangluan Xu and
                  Xian Sun and
                  Kun Fu},
  title        = {NavAgent: Multi-scale Urban Street View Fusion For {UAV} Embodied
                  Vision-and-Language Navigation},
  journal      = {CoRR},
  volume       = {abs/2411.08579},
  year         = {2024},
  doi          = {10.48550/ARXIV.2411.08579},
  eprinttype    = {arXiv},
  eprint       = {2411.08579},
  bibsource    = {dblp computer science bibliography, https://dblp.org}
}

@article{flightgpt,
  author       = {Hengxing Cai and
                  Jinhan Dong and
                  Jingjun Tan and
                  Jingcheng Deng and
                  Sihang Li and
                  Zhifeng Gao and
                  Haidong Wang and
                  Zicheng Su and
                  Agachai Sumalee and
                  Renxin Zhong},
  title        = {FlightGPT: Towards Generalizable and Interpretable {UAV} Vision-and-Language
                  Navigation with Vision-Language Models},
  journal      = {CoRR},
  volume       = {abs/2505.12835},
  year         = {2025},
  doi          = {10.48550/ARXIV.2505.12835},
  eprinttype    = {arXiv},
  eprint       = {2505.12835},
  bibsource    = {dblp computer science bibliography, https://dblp.org}
}

@inproceedings{citynavagent,
  author       = {Weichen Zhang and
                  Chen Gao and
                  Shiquan Yu and
                  Ruiying Peng and
                  Baining Zhao and
                  Qian Zhang and
                  Jinqiang Cui and
                  Xinlei Chen and
                  Yong Li},
  editor       = {Wanxiang Che and
                  Joyce Nabende and
                  Ekaterina Shutova and
                  Mohammad Taher Pilehvar},
  title        = {CityNavAgent: Aerial Vision-and-Language Navigation with Hierarchical
                  Semantic Planning and Global Memory},
  booktitle    = {Proceedings of the 63rd Annual Meeting of the Association for Computational
                  Linguistics (Volume 1: Long Papers), {ACL} 2025, Vienna, Austria,
                  July 27 - August 1, 2025},
  pages        = {31292--31309},
  publisher    = {Association for Computational Linguistics},
  year         = {2025},
  bibsource    = {dblp computer science bibliography, https://dblp.org}
}

@article{qwen3vl,
  author       = {Shuai Bai and
                  Yuxuan Cai and
                  Ruizhe Chen and
                  Keqin Chen and
                  Xionghui Chen and
                  Zesen Cheng and
                  Lianghao Deng and
                  Wei Ding and
                  Chang Gao and
                  Chunjiang Ge and
                  Wenbin Ge and
                  Zhifang Guo and
                  Qidong Huang and
                  Jie Huang and
                  Fei Huang and
                  Binyuan Hui and
                  Shutong Jiang and
                  Zhaohai Li and
                  Mingsheng Li and
                  Mei Li and
                  Kaixin Li and
                  Zicheng Lin and
                  Junyang Lin and
                  Xuejing Liu and
                  Jiawei Liu and
                  Chenglong Liu and
                  Yang Liu and
                  Dayiheng Liu and
                  Shixuan Liu and
                  Dunjie Lu and
                  Ruilin Luo and
                  Chenxu Lv and
                  Rui Men and
                  Lingchen Meng and
                  Xuancheng Ren and
                  Xingzhang Ren and
                  Sibo Song and
                  Yuchong Sun and
                  Jun Tang and
                  Jianhong Tu and
                  Jianqiang Wan and
                  Peng Wang and
                  Pengfei Wang and
                  Qiuyue Wang and
                  Yuxuan Wang and
                  Tianbao Xie and
                  Yiheng Xu and
                  Haiyang Xu and
                  Jin Xu and
                  Zhibo Yang and
                  Mingkun Yang and
                  Jianxin Yang and
                  An Yang and
                  Bowen Yu and
                  Fei Zhang and
                  Hang Zhang and
                  Xi Zhang and
                  Bo Zheng and
                  Humen Zhong and
                  Jingren Zhou and
                  Fan Zhou and
                  Jing Zhou and
                  Yuanzhi Zhu and
                  Ke Zhu},
  title        = {Qwen3-VL Technical Report},
  journal      = {CoRR},
  volume       = {abs/2511.21631},
  year         = {2025},
  doi          = {10.48550/ARXIV.2511.21631},
  eprinttype    = {arXiv},
  eprint       = {2511.21631},
  bibsource    = {dblp computer science bibliography, https://dblp.org}
}

@inproceedings{frontier,
  author       = {Brian Yamauchi},
  title        = {A frontier-based approach for autonomous exploration},
  booktitle    = {Proceedings 1997 {IEEE} International Symposium on Computational Intelligence
                  in Robotics and Automation CIRA'97 - Towards New Computational Principles
                  for Robotics and Automation, July 10-11, 1997, Monterey, California,
                  {USA}},
  pages        = {146--151},
  publisher    = {{IEEE} Computer Society},
  year         = {1997},
  doi          = {10.1109/CIRA.1997.613851},
  bibsource    = {dblp computer science bibliography, https://dblp.org}
}

@inproceedings{l3mvn,
  author       = {Bangguo Yu and
                  Hamidreza Kasaei and
                  Ming Cao},
  title        = {{L3MVN:} Leveraging Large Language Models for Visual Target Navigation},
  booktitle    = {{IROS}},
  pages        = {3554--3560},
  year         = {2023},
  doi          = {10.1109/IROS55552.2023.10342512},
  bibsource    = {dblp computer science bibliography, https://dblp.org}
}

@article{citynav,
  author       = {Jungdae Lee and
                  Taiki Miyanishi and
                  Shuhei Kurita and
                  Koya Sakamoto and
                  Daichi Azuma and
                  Yutaka Matsuo and
                  Nakamasa Inoue},
  title        = {CityNav: Language-Goal Aerial Navigation Dataset with Geographic Information},
  journal      = {CoRR},
  volume       = {abs/2406.14240},
  year         = {2024},
  doi          = {10.48550/ARXIV.2406.14240},
  eprinttype    = {arXiv},
  eprint       = {2406.14240},
  bibsource    = {dblp computer science bibliography, https://dblp.org}
}

@article{vlfly,
  author       = {Yuhang Zhang and
                  Haosheng Yu and
                  Jiaping Xiao and
                  Mir Feroskhan},
  title        = {Grounded Vision-Language Navigation for UAVs with Open-Vocabulary Goal Understanding},
  journal      = {CoRR},
  volume       = {abs/2506.10756},
  year         = {2025},
  doi          = {10.48550/ARXIV.2506.10756},
  eprinttype    = {arXiv},
  eprint       = {2506.10756},
  bibsource    = {dblp computer science bibliography, https://dblp.org}
}

@article{openfly,
  author       = {Yunpeng Gao and
                  Chenhui Li and
                  Zhongrui You and
                  Junli Liu and
                  Zhen Li and
                  Pengan Chen and
                  Qizhi Chen and
                  Zhonghan Tang and
                  Liansheng Wang and
                  Penghui Yang and
                  Yiwen Tang and
                  Yuhang Tang and
                  Shuai Liang and
                  Songyi Zhu and
                  Ziqin Xiong and
                  Yifei Su and
                  Xinyi Ye and
                  Jianan Li and
                  Yan Ding and
                  Dong Wang and
                  Zhigang Wang and
                  Bin Zhao and
                  Xuelong Li},
  title        = {OpenFly: {A} Versatile Toolchain and Large-scale Benchmark for Aerial
                  Vision-Language Navigation},
  journal      = {CoRR},
  volume       = {abs/2502.18041},
  year         = {2025},
  doi          = {10.48550/ARXIV.2502.18041},
  eprinttype    = {arXiv},
  eprint       = {2502.18041},
  bibsource    = {dblp computer science bibliography, https://dblp.org}
}

@inproceedings{fela,
  author       = {Yifei Su and
                  Dong An and
                  Kehan Chen and
                  Weichen Yu and
                  Baiyang Ning and
                  Yonggen Ling and
                  Yan Huang and
                  Liang Wang},
  editor       = {Toby Walsh and
                  Julie Shah and
                  Zico Kolter},
  title        = {Learning Fine-Grained Alignment for Aerial Vision-Dialog Navigation},
  booktitle    = {AAAI-25, Sponsored by the Association for the Advancement of Artificial
                  Intelligence, February 25 - March 4, 2025, Philadelphia, PA, {USA}},
  pages        = {7060--7068},
  publisher    = {{AAAI} Press},
  year         = {2025},
  doi          = {10.1609/AAAI.V39I7.32758},
  bibsource    = {dblp computer science bibliography, https://dblp.org}
}

@article{Fanglong,
author = {Fanglong Yao and Youzhi Liu and Wenyi Zhang and Zhengqiu Zhu and Chenglong Li and Nayu Liu and Peng Hu and Yuanchang Yue and Kaiwen Wei and Xin He and Xudong Zhao and Zihan Wei and Haotian Xu and Zhiyuan Wang and Gujie Shao and Liu Yang and Dan Zhao and Yong Yang},
title = {AeroVerse-Review: Comprehensive survey on aerial embodied vision-and-language navigation},
journal = {The Innovation Informatics},
volume = {1},
number = {1},
pages = {100015},
year = {2025},
issn = {3105-8515},
doi = {10.59717/j.xinn-inform.2025.100015}
}

@article{vlmsurvey,
  author       = {Jingyi Zhang and
                  Jiaxing Huang and
                  Sheng Jin and
                  Shijian Lu},
  title        = {Vision-Language Models for Vision Tasks: {A} Survey},
  journal      = {{IEEE} Trans. Pattern Anal. Mach. Intell.},
  volume       = {46},
  number       = {8},
  pages        = {5625--5644},
  year         = {2024},
  doi          = {10.1109/TPAMI.2024.3369699},
  bibsource    = {dblp computer science bibliography, https://dblp.org}
}

@article{uavai,
  author       = {Ranjan Sapkota and
                  Konstantinos I. Roumeliotis and
                  Manoj Karkee},
  title        = {UAVs Meet Agentic {AI:} {A} Multidomain Survey of Autonomous Aerial
                  Intelligence and Agentic UAVs},
  journal      = {CoRR},
  volume       = {abs/2506.08045},
  year         = {2025},
  doi          = {10.48550/ARXIV.2506.08045},
  eprinttype    = {arXiv},
  eprint       = {2506.08045},
  bibsource    = {dblp computer science bibliography, https://dblp.org}
}

@inproceedings{geotext,
  author       = {Meng Chu and
                  Zhedong Zheng and
                  Wei Ji and
                  Tingyu Wang and
                  Tat{-}Seng Chua},
  editor       = {Ales Leonardis and
                  Elisa Ricci and
                  Stefan Roth and
                  Olga Russakovsky and
                  Torsten Sattler and
                  G{\"{u}}l Varol},
  title        = {Towards Natural Language-Guided Drones: GeoText-1652 Benchmark with
                  Spatial Relation Matching},
  booktitle    = {Computer Vision - {ECCV} 2024 - 18th European Conference, Milan, Italy,
                  September 29-October 4, 2024, Proceedings, Part {XI}},
  series       = {Lecture Notes in Computer Science},
  volume       = {15069},
  pages        = {213--231},
  publisher    = {Springer},
  year         = {2024},
  doi          = {10.1007/978-3-031-73247-8\_13},
  bibsource    = {dblp computer science bibliography, https://dblp.org}
}

@article{aeroduo,
  author       = {Ruipu Wu and
                  Yige Zhang and
                  Jinyu Chen and
                  Linjiang Huang and
                  Shifeng Zhang and
                  Xu Zhou and
                  Liang Wang and
                  Si Liu},
  title        = {AeroDuo: Aerial Duo for UAV-based Vision and Language Navigation},
  journal      = {CoRR},
  volume       = {abs/2508.15232},
  year         = {2025},
  doi          = {10.48550/ARXIV.2508.15232},
  eprinttype    = {arXiv},
  eprint       = {2508.15232},
  bibsource    = {dblp computer science bibliography, https://dblp.org}
}

@article{generalpurpose,
  author       = {Ji Zhao and
                  Xiao Lin},
  title        = {General-Purpose Aerial Intelligent Agents Empowered by Large Language
                  Models},
  journal      = {CoRR},
  volume       = {abs/2503.08302},
  year         = {2025},
  doi          = {10.48550/ARXIV.2503.08302},
  eprinttype    = {arXiv},
  eprint       = {2503.08302},
  bibsource    = {dblp computer science bibliography, https://dblp.org}
}

@article{rlmethod,
  author       = {Shijin Zhao and
                  Fuhui Zhou and
                  Qihui Wu},
  title        = {{AAV} Visual Navigation in the Large-Scale Outdoor Environment: {A}
                  Semantic-Map-Based Cognitive Escape Reinforcement Learning Method},
  journal      = {{IEEE} Internet Things J.},
  volume       = {12},
  number       = {11},
  pages        = {15926--15938},
  year         = {2025},
  doi          = {10.1109/JIOT.2025.3532164},
  bibsource    = {dblp computer science bibliography, https://dblp.org}
}

@article{airvista,
  author       = {Fei Lin and
                  Yonglin Tian and
                  Tengchao Zhang and
                  Jun Huang and
                  Sangtian Guan and
                  Fei{-}Yue Wang},
  title        = {AirVista-II: An Agentic System for Embodied UAVs Toward Dynamic Scene
                  Semantic Understanding},
  journal      = {CoRR},
  volume       = {abs/2504.09583},
  year         = {2025},
  doi          = {10.48550/ARXIV.2504.09583},
  eprinttype    = {arXiv},
  eprint       = {2504.09583},
  bibsource    = {dblp computer science bibliography, https://dblp.org}
}

@article{refdrone,
  author       = {Zhichao Sun and
                  Yepeng Liu and
                  Huachao Zhu and
                  Yuliang Gu and
                  Yuda Zou and
                  Zelong Liu and
                  Gui{-}Song Xia and
                  Bo Du and
                  Yongchao Xu},
  title        = {RefDrone: {A} Challenging Benchmark for Referring Expression Comprehension
                  in Drone Scenes},
  journal      = {CoRR},
  volume       = {abs/2502.00392},
  year         = {2025},
  doi          = {10.48550/ARXIV.2502.00392},
  eprinttype    = {arXiv},
  eprint       = {2502.00392},
  bibsource    = {dblp computer science bibliography, https://dblp.org}
}

@inproceedings{selfprompting,
  author       = {Nianxin Li and
                  Mao Ye and
                  Lihua Zhou and
                  Song Tang and
                  Yan Gan and
                  Zizhuo Liang and
                  Xiatian Zhu},
  editor       = {Toby Walsh and
                  Julie Shah and
                  Zico Kolter},
  title        = {Self-Prompting Analogical Reasoning for {UAV} Object Detection},
  booktitle    = {AAAI-25, Sponsored by the Association for the Advancement of Artificial
                  Intelligence, February 25 - March 4, 2025, Philadelphia, PA, {USA}},
  pages        = {18412--18420},
  publisher    = {{AAAI} Press},
  year         = {2025},
  doi          = {10.1609/AAAI.V39I17.34026},
  bibsource    = {dblp computer science bibliography, https://dblp.org}
}

@article{geonav,
  author       = {Haotian Xu and
                  Yue Hu and
                  Chen Gao and
                  Zhengqiu Zhu and
                  Yong Zhao and
                  Yong Li and
                  Quanjun Yin},
  title        = {GeoNav: Empowering MLLMs with Explicit Geospatial Reasoning Abilities
                  for Language-Goal Aerial Navigation},
  journal      = {CoRR},
  volume       = {abs/2504.09587},
  year         = {2025},
  doi          = {10.48550/ARXIV.2504.09587},
  eprinttype    = {arXiv},
  eprint       = {2504.09587},
  bibsource    = {dblp computer science bibliography, https://dblp.org}
}

@article{avlngrid,
  author       = {Ganlong Zhao and
                  Guanbin Li and
                  Jia Pan and
                  Yizhou Yu},
  title        = {Aerial Vision-and-Language Navigation with Grid-based View Selection
                  and Map Construction},
  journal      = {CoRR},
  volume       = {abs/2503.11091},
  year         = {2025},
  doi          = {10.48550/ARXIV.2503.11091},
  eprinttype    = {arXiv},
  eprint       = {2503.11091},
  bibsource    = {dblp computer science bibliography, https://dblp.org}
}

@article{uavllm,
  author       = {Yonglin Tian and
                  Fei Lin and
                  Yiduo Li and
                  Tengchao Zhang and
                  Qiyao Zhang and
                  Xuan Fu and
                  Jun Huang and
                  Xingyuan Dai and
                  Yutong Wang and
                  Chunwei Tian and
                  Bai Li and
                  Yisheng Lv and
                  Levente Kov{\'{a}}cs and
                  Feiyue Wang},
  title        = {UAVs meet LLMs: Overviews and perspectives towards agentic low-altitude
                  mobility},
  journal      = {Inf. Fusion},
  volume       = {122},
  pages        = {103158},
  year         = {2025},
  doi          = {10.1016/J.INFFUS.2025.103158},
  bibsource    = {dblp computer science bibliography, https://dblp.org}
}

@article{aeroverse,
  author       = {Fanglong Yao and
                  Yuanchang Yue and
                  Youzhi Liu and
                  Xian Sun and
                  Kun Fu},
  title        = {AeroVerse: UAV-Agent Benchmark Suite for Simulating, Pre-training,
                  Finetuning, and Evaluating Aerospace Embodied World Models},
  journal      = {CoRR},
  volume       = {abs/2408.15511},
  year         = {2024},
  doi          = {10.48550/ARXIV.2408.15511},
  eprinttype    = {arXiv},
  eprint       = {2408.15511},
  bibsource    = {dblp computer science bibliography, https://dblp.org}
}

@article{racevla,
  author       = {Valerii Serpiva and
                  Artem Lykov and
                  Artyom Myshlyaev and
                  Muhammad Haris Khan and
                  Ali Alridha Abdulkarim and
                  Oleg Sautenkov and
                  Dzmitry Tsetserukou},
  title        = {RaceVLA: VLA-based Racing Drone Navigation with Human-like Behaviour},
  journal      = {CoRR},
  volume       = {abs/2503.02572},
  year         = {2025},
  doi          = {10.48550/ARXIV.2503.02572},
  eprinttype    = {arXiv},
  eprint       = {2503.02572},
  bibsource    = {dblp computer science bibliography, https://dblp.org}
}

@article{cognitivedrone,
  author       = {Artem Lykov and
                  Valerii Serpiva and
                  Muhammad Haris Khan and
                  Oleg Sautenkov and
                  Artyom Myshlyaev and
                  Grik Tadevosyan and
                  Yasheerah Yaqoot and
                  Dzmitry Tsetserukou},
  title        = {CognitiveDrone: {A} {VLA} Model and Evaluation Benchmark for Real-Time
                  Cognitive Task Solving and Reasoning in UAVs},
  journal      = {CoRR},
  volume       = {abs/2503.01378},
  year         = {2025},
  doi          = {10.48550/ARXIV.2503.01378},
  eprinttype    = {arXiv},
  eprint       = {2503.01378},
  bibsource    = {dblp computer science bibliography, https://dblp.org}
}

@article{uavflow,
  author       = {Xiangyu Wang and
                  Donglin Yang and
                  Yue Liao and
                  Wenhao Zheng and
                  Wenjun Wu and
                  Bin Dai and
                  Hongsheng Li and
                  Si Liu},
  title        = {UAV-Flow Colosseo: {A} Real-World Benchmark for Flying-on-a-Word {UAV}
                  Imitation Learning},
  journal      = {CoRR},
  volume       = {abs/2505.15725},
  year         = {2025},
  doi          = {10.48550/ARXIV.2505.15725},
  eprinttype    = {arXiv},
  eprint       = {2505.15725},
  bibsource    = {dblp computer science bibliography, https://dblp.org}
}

@article{raven,
  author       = {Seungchan Kim and
                  Omar Alama and
                  Dmytro Kurdydyk and
                  John Keller and
                  Nikhil Varma Keetha and
                  Wenshan Wang and
                  Yonatan Bisk and
                  Sebastian A. Scherer},
  title        = {{RAVEN:} Resilient Aerial Navigation via Open-Set Semantic Memory
                  and Behavior Adaptation},
  journal      = {CoRR},
  volume       = {abs/2509.23563},
  year         = {2025},
  doi          = {10.48550/ARXIV.2509.23563},
  eprinttype    = {arXiv},
  eprint       = {2509.23563},
  bibsource    = {dblp computer science bibliography, https://dblp.org}
}

@article{rapid,
  author       = {Minwoo Kim and
                  Geun Sik Bae and
                  Jinwoo Lee and
                  Woojae Shin and
                  Changseung Kim and
                  Myong{-}Yol Choi and
                  Heejung Shin and
                  Hyondong Oh},
  title        = {{RAPID:} Robust and Agile Planner Using Inverse Reinforcement Learning
                  for Vision-Based Drone Navigation},
  journal      = {CoRR},
  volume       = {abs/2502.02054},
  year         = {2025},
  doi          = {10.48550/ARXIV.2502.02054},
  eprinttype    = {arXiv},
  eprint       = {2502.02054},
  bibsource    = {dblp computer science bibliography, https://dblp.org}
}

@article{vlnce,
  author       = {Kehan Chen and
                  Dong An and
                  Yan Huang and
                  Rongtao Xu and
                  Yifei Su and
                  Yonggen Ling and
                  Ian D. Reid and
                  Liang Wang},
  title        = {Constraint-Aware Zero-Shot Vision-Language Navigation in Continuous
                  Environments},
  journal      = {{IEEE} Trans. Pattern Anal. Mach. Intell.},
  volume       = {47},
  number       = {11},
  pages        = {10441--10456},
  year         = {2025},
  doi          = {10.1109/TPAMI.2025.3594204},
  bibsource    = {dblp computer science bibliography, https://dblp.org}
}

@article{mmuavbench,
  author       = {Shiqi Dai and
                  Zizhi Ma and
                  Zhicong Luo and
                  Xuesong Yang and
                  Yibin Huang and
                  Wanyue Zhang and
                  Chi Chen and
                  Zonghao Guo and
                  Wang Xu and
                  Yufei Sun and
                  Maosong Sun},
  title        = {MM-UAVBench: How Well Do Multimodal Large Language Models See, Think,
                  and Plan in Low-Altitude {UAV} Scenarios?},
  journal      = {CoRR},
  volume       = {abs/2512.23219},
  year         = {2025},
  doi          = {10.48550/ARXIV.2512.23219},
  eprinttype    = {arXiv},
  eprint       = {2512.23219},
  bibsource    = {dblp computer science bibliography, https://dblp.org}
}

@article{uavbench,
  author       = {Mohamed Amine Ferrag and
                  Abderrahmane Lakas and
                  M{\'{e}}rouane Debbah},
  title        = {UAVBench: An Open Benchmark Dataset for Autonomous and Agentic {AI}
                  {UAV} Systems via LLM-Generated Flight Scenarios},
  journal      = {CoRR},
  volume       = {abs/2511.11252},
  year         = {2025},
  doi          = {10.48550/ARXIV.2511.11252},
  eprinttype    = {arXiv},
  eprint       = {2511.11252},
  bibsource    = {dblp computer science bibliography, https://dblp.org}
}

@article{ster,
  author       = {Huilin Xu and
                  Zhuoyang Liu and
                  Yixiang Luomei and
                  Feng Xu},
  title        = {Aerial Vision-Language Navigation with a Unified Framework for Spatial,
                  Temporal and Embodied Reasoning},
  journal      = {CoRR},
  volume       = {abs/2512.08639},
  year         = {2025},
  doi          = {10.48550/ARXIV.2512.08639},
  eprinttype    = {arXiv},
  eprint       = {2512.08639},
  bibsource    = {dblp computer science bibliography, https://dblp.org}
}

@article{vlaan,
  author       = {Yuze Wu and
                  Mo Zhu and
                  Xingxing Li and
                  Yuheng Du and
                  Yuxin Fan and
                  Wenjun Li and
                  Zhichao Han and
                  Xin Zhou and
                  Fei Gao},
  title        = {{VLA-AN:} An Efficient and Onboard Vision-Language-Action Framework
                  for Aerial Navigation in Complex Environments},
  journal      = {CoRR},
  volume       = {abs/2512.15258},
  year         = {2025},
  doi          = {10.48550/ARXIV.2512.15258},
  eprinttype    = {arXiv},
  eprint       = {2512.15258},
  bibsource    = {dblp computer science bibliography, https://dblp.org}
}

@article{indooruav,
  author       = {Xu Liu and
                  Yu Liu and
                  Hanshuo Qiu and
                  Qirong Yang and
                  Zhouhui Lian},
  title        = {IndoorUAV: Benchmarking Vision-Language {UAV} Navigation in Continuous
                  Indoor Environments},
  journal      = {CoRR},
  volume       = {abs/2512.19024},
  year         = {2025},
  doi          = {10.48550/ARXIV.2512.19024},
  eprinttype    = {arXiv},
  eprint       = {2512.19024},
  bibsource    = {dblp computer science bibliography, https://dblp.org}
}

@article{longfly,
  author       = {Wen Jiang and
                  Li Wang and
                  Kangyao Huang and
                  Wei Fan and
                  Jinyuan Liu and
                  Shaoyu Liu and
                  Hongwei Duan and
                  Bin Xu and
                  Xiangyang Ji},
  title        = {LongFly: Long-Horizon {UAV} Vision-and-Language Navigation with Spatiotemporal
                  Context Integration},
  journal      = {CoRR},
  volume       = {abs/2512.22010},
  year         = {2025},
  doi          = {10.48550/ARXIV.2512.22010},
  eprinttype    = {arXiv},
  eprint       = {2512.22010},
  bibsource    = {dblp computer science bibliography, https://dblp.org}
}

@article{uavvlrr,
  author       = {Yasheerah Yaqoot and
                  Muhammad Ahsan Mustafa and
                  Oleg Sautenkov and
                  Dzmitry Tsetserukou},
  title        = {{UAV-VLRR:} Vision-Language Informed {NMPC} for Rapid Response in
                  {UAV} Search and Rescue},
  journal      = {CoRR},
  volume       = {abs/2503.02465},
  year         = {2025},
  doi          = {10.48550/ARXIV.2503.02465},
  eprinttype    = {arXiv},
  eprint       = {2503.02465},
  bibsource    = {dblp computer science bibliography, https://dblp.org}
}

@article{hett,
  author       = {Xichen Ding and
                  Jianzhe Gao and
                  Cong Pan and
                  Wenguan Wang and
                  Jie Qin},
  title        = {History-Enhanced Two-Stage Transformer for Aerial Vision-and-Language
                  Navigation},
  journal      = {CoRR},
  volume       = {abs/2512.14222},
  year         = {2025},
  doi          = {10.48550/ARXIV.2512.14222},
  eprinttype    = {arXiv},
  eprint       = {2512.14222},
  bibsource    = {dblp computer science bibliography, https://dblp.org}
}

@article{spf,
  author       = {Chih Yao Hu and
                  Yang{-}Sen Lin and
                  Yuna Lee and
                  Chih{-}Hai Su and
                  Jie{-}Ying Lee and
                  Shr{-}Ruei Tsai and
                  Chin{-}Yang Lin and
                  Kuan{-}Wen Chen and
                  Tsung{-}Wei Ke and
                  Yu{-}Lun Liu},
  title        = {See, Point, Fly: {A} Learning-Free {VLM} Framework for Universal Unmanned
                  Aerial Navigation},
  journal      = {CoRR},
  volume       = {abs/2509.22653},
  year         = {2025},
  doi          = {10.48550/ARXIV.2509.22653},
  eprinttype    = {arXiv},
  eprint       = {2509.22653},
  bibsource    = {dblp computer science bibliography, https://dblp.org}
}

@article{airnav,
  author       = {Hengxing Cai and
                  Yijie Rao and
                  Ligang Huang and
                  Zanyang Zhong and
                  Jinhan Dong and
                  Jingjun Tan and
                  Wenhao Lu and
                  Renxin Zhong},
  title        = {AirNav: {A} Large-Scale Real-World {UAV} Vision-and-Language Navigation
                  Dataset with Natural and Diverse Instructions},
  journal      = {CoRR},
  volume       = {abs/2601.03707},
  year         = {2026},
  doi          = {10.48550/ARXIV.2601.03707},
  eprinttype    = {arXiv},
  eprint       = {2601.03707},
  bibsource    = {dblp computer science bibliography, https://dblp.org}
}

@inproceedings{airsim,
  author       = {Shital Shah and
                  Debadeepta Dey and
                  Chris Lovett and
                  Ashish Kapoor},
  editor       = {Marco Hutter and
                  Roland Siegwart},
  title        = {AirSim: High-Fidelity Visual and Physical Simulation for Autonomous
                  Vehicles},
  booktitle    = {Field and Service Robotics, Results of the 11th International Conference,
                  {FSR} 2017, Zurich, Switzerland, 12-15 September 2017},
  series       = {Springer Proceedings in Advanced Robotics},
  volume       = {5},
  pages        = {621--635},
  publisher    = {Springer},
  year         = {2017},
  doi          = {10.1007/978-3-319-67361-5\_40},
  bibsource    = {dblp computer science bibliography, https://dblp.org}
}

@inproceedings{yoloworld,
  author       = {Tianheng Cheng and
                  Lin Song and
                  Yixiao Ge and
                  Wenyu Liu and
                  Xinggang Wang and
                  Ying Shan},
  title        = {YOLO-World: Real-Time Open-Vocabulary Object Detection},
  booktitle    = {{IEEE/CVF} Conference on Computer Vision and Pattern Recognition,
                  {CVPR} 2024, Seattle, WA, USA, June 16-22, 2024},
  pages        = {16901--16911},
  publisher    = {{IEEE}},
  year         = {2024},
  doi          = {10.1109/CVPR52733.2024.01599},
  bibsource    = {dblp computer science bibliography, https://dblp.org}
}

@article{cnn,
  author       = {Luka Siktar and
                  Branimir Caran and
                  Bojan Sekoranja and
                  Marko Svaco},
  title        = {Autonomous {UAV} Navigation for Search and Rescue Missions Using Computer
                  Vision and Convolutional Neural Networks},
  journal      = {CoRR},
  volume       = {abs/2507.18160},
  year         = {2025},
  doi          = {10.48550/ARXIV.2507.18160},
  eprinttype    = {arXiv},
  eprint       = {2507.18160},
  bibsource    = {dblp computer science bibliography, https://dblp.org}
}

@article{drespnet,
  author       = {Aykut Sirma and
                  Angelos Plastropoulos and
                  Argyrios Zolotas and
                  Gilbert Tang},
  title        = {DRespNeT: {A} {UAV} Dataset and YOLOv8-DRN Model for Aerial Instance
                  Segmentation of Building Access Points for Post-Earthquake Search-and-Rescue
                  Missions},
  journal      = {CoRR},
  volume       = {abs/2508.16016},
  year         = {2025},
  doi          = {10.48550/ARXIV.2508.16016},
  eprinttype    = {arXiv},
  eprint       = {2508.16016},
  bibsource    = {dblp computer science bibliography, https://dblp.org}
}

@article{yolov10,
  author       = {Aditya Mishra and
                  Modigari Narendra and
                  Aryan Sinha and
                  Aakash Kumar},
  title        = {Dynamic Backbone Optimization of YOLOv10 for Real-Time Object Detection
                  in UAV-Based Search and Rescue Missions},
  journal      = {{IEEE} Access},
  volume       = {13},
  pages        = {195975--195986},
  year         = {2025},
  doi          = {10.1109/ACCESS.2025.3633379},
  bibsource    = {dblp computer science bibliography, https://dblp.org}
}

@article{realexp,
  author       = {Stella Dumencic and
                  Luka Lanca and
                  Karlo Jakac and
                  Stefan Ivic},
  title        = {Experimental validation of {UAV} search and detection system in real
                  wilderness environment},
  journal      = {CoRR},
  volume       = {abs/2502.17372},
  year         = {2025},
  doi          = {10.48550/ARXIV.2502.17372},
  eprinttype    = {arXiv},
  eprint       = {2502.17372},
  bibsource    = {dblp computer science bibliography, https://dblp.org}
}

@article{smartagent,
  author       = {Zijian Ge and
                  Jingjing Jiang and
                  Matthew Coombes},
  title        = {Multi-UAV Search and Rescue in Wilderness Using Smart Agent-Based
                  Probability Models},
  journal      = {CoRR},
  volume       = {abs/2411.10148},
  year         = {2024},
  doi          = {10.48550/ARXIV.2411.10148},
  eprinttype    = {arXiv},
  eprint       = {2411.10148},
  bibsource    = {dblp computer science bibliography, https://dblp.org}
}

@article{cognitive,
  author       = {Jane Cleland{-}Huang and
                  Pedro Alarcon Granadeno and
                  Arturo Miguel Russell Bernal and
                  Demetrius Hernandez and
                  Michael Murphy and
                  Maureen Petterson and
                  Walter J. Scheirer},
  title        = {Cognitive Guardrails for Open-World Decision Making in Autonomous
                  Drone Swarms},
  journal      = {CoRR},
  volume       = {abs/2505.23576},
  year         = {2025},
  doi          = {10.48550/ARXIV.2505.23576},
  eprinttype    = {arXiv},
  eprint       = {2505.23576},
  bibsource    = {dblp computer science bibliography, https://dblp.org}
}

@article{nbv,
  author       = {Sigrid Strand and
                  Thomas Wiedemann and
                  Bram Burczek and
                  Dmitriy Shutin},
  title        = {Enhancing {UAV} Search Under Occlusion Using Next Best View Planning},
  journal      = {{IEEE} J. Sel. Top. Appl. Earth Obs. Remote. Sens.},
  volume       = {19},
  pages        = {1085--1096},
  year         = {2026},
  doi          = {10.1109/JSTARS.2025.3638881},
  bibsource    = {dblp computer science bibliography, https://dblp.org}
}

@article{dual,
  author       = {Guang Yang and
                  Yadong Mo and
                  Chengyu Lv and
                  Ying Zhang and
                  Jian Li and
                  Shimin Wei},
  title        = {A dual-layer task planning algorithm based on UAVs-human cooperation
                  for search and rescue},
  journal      = {Appl. Soft Comput.},
  volume       = {181},
  pages        = {113488},
  year         = {2025},
  doi          = {10.1016/J.ASOC.2025.113488},
  bibsource    = {dblp computer science bibliography, https://dblp.org}
}

@article{drl,
  author       = {Thomas Hickling and
                  Maxwell Hogan and
                  Abdulla Tammam and
                  Nabil Aouf},
  title        = {Deep Reinforcement Learning based Autonomous Decision-Making for Cooperative
                  UAVs: {A} Search and Rescue Real World Application},
  journal      = {CoRR},
  volume       = {abs/2502.20326},
  year         = {2025},
  doi          = {10.48550/ARXIV.2502.20326},
  eprinttype    = {arXiv},
  eprint       = {2502.20326},
  bibsource    = {dblp computer science bibliography, https://dblp.org}
}

@inproceedings{coverpath,
  author       = {Julian Bialas and
                  Mario D{\"{o}}ller and
                  Simone Walch and
                  Michiel Jan Van Veelen and
                  Abraham Mejia{-}Aguilar},
  title        = {Optimizing Multi-Agent Coverage Path Planning {UAV} Search and Rescue
                  Missions with Prioritizing Deep Reinforcement Learning},
  booktitle    = {{IEEE} International Conference on Robotics and Biomimetics, {ROBIO}
                  2024, Bangkok, Thailand, December 10-14, 2024},
  pages        = {85--90},
  publisher    = {{IEEE}},
  year         = {2024},
  doi          = {10.1109/ROBIO64047.2024.10907496},
  bibsource    = {dblp computer science bibliography, https://dblp.org}
}

@inproceedings{llmcollabration,
  author       = {M. I. R. Shuvo and
                  Nafiul Alam and
                  Awal Ahmed Fime and
                  Hannah Lee and
                  Xiangxu Lin and
                  Jong{-}Hoon Kim},
  title        = {A Novel Large Language Model {(LLM)} Based Approach for Robotic Collaboration
                  in Search and Rescue Operations},
  booktitle    = {50th Annual Conference of the {IEEE} Industrial Electronics Society,
                  {IECON} 2024, Chicago, IL, USA, November 3-6, 2024},
  pages        = {1--6},
  publisher    = {{IEEE}},
  year         = {2024},
  doi          = {10.1109/IECON55916.2024.10905094},
  bibsource    = {dblp computer science bibliography, https://dblp.org}
}

@article{llmsar,
  author       = {Kailun Ji and
                  Xiaoyu Hu and
                  Xinyu Zhang and
                  Jun Chen},
  title        = {An LLM-based Framework for Human-Swarm Teaming Cognition in Disaster
                  Search and Rescue},
  journal      = {CoRR},
  volume       = {abs/2511.04042},
  year         = {2025},
  doi          = {10.48550/ARXIV.2511.04042},
  eprinttype    = {arXiv},
  eprint       = {2511.04042},
  bibsource    = {dblp computer science bibliography, https://dblp.org}
}

@inproceedings{llmselect,
  author       = {Dimitrios Panagopoulos and
                  Adolfo Perrusqu{\'{\i}}a and
                  Weisi Guo},
  title        = {Selective Exploration and Information Gathering in Search and Rescue
                  Using Hierarchical Learning Guided by Natural Language Input},
  booktitle    = {{IEEE} International Conference on Systems, Man, and Cybernetics,
                  {SMC} 2024, Kuching, Malaysia, October 6-10, 2024},
  pages        = {1175--1180},
  publisher    = {{IEEE}},
  year         = {2024},
  doi          = {10.1109/SMC54092.2024.10831125},
  bibsource    = {dblp computer science bibliography, https://dblp.org}
}

@article{AUSPEX,
  author       = {Bjorn D{\"{o}}schl and
                  Kai Sommer and
                  Jane Jean Kiam},
  title        = {{AUSPEX:} An integrated open-source decision-making framework for
                  UAVs in rescue missions},
  journal      = {Frontiers Robotics {AI}},
  volume       = {12},
  year         = {2025},
  doi          = {10.3389/FROBT.2025.1583479},
  bibsource    = {dblp computer science bibliography, https://dblp.org}
}

@inproceedings{SaynFly,
  author       = {Bjorn D{\"{o}}schl and
                  Jane Jean Kiam},
  title        = {Say'n'Fly: An LLM-Modulo Online Planning Framework to Automate
                  {UAV} Command and Control},
  booktitle    = {34th {IEEE} International Conference on Robot and Human Interactive
                  Communication, {RO-MAN} 2025, Eindhoven, Netherlands, August 25-29,
                  2025},
  pages        = {1693--1698},
  publisher    = {{IEEE}},
  year         = {2025},
  doi          = {10.1109/RO-MAN63969.2025.11217764},
  bibsource    = {dblp computer science bibliography, https://dblp.org}
}

\newpage
\appendix

\section{Quantitative Difficulty Scoring}
\label{sec:difficulty}
\begin{table}[htbp]
    \centering
    \caption{Quantitative Difficulty Scoring Matrix for Embodied SAR Tasks.}
    \renewcommand{\arraystretch}{1.2}
    \begin{tabularx}{0.98\textwidth}{@{}Xl@{\hspace{2em}}c@{}}
        \toprule
        \textbf{Dimension} & \textbf{Condition} & \textbf{Score} \\ 
        \midrule
        
        \multirow{4}{*}{A. Average Distance} 
        & $d \le 116.6$ m & $+1$ \\
        & $116.6 < d \le 230.3$ m & $+2$ \\
        & $230.3 < d \le 373.6$ m & $+3$ \\
        & $d > 373.6$ m & $+4$ \\ 
        \midrule
        
        \multirow{3}{*}{B. Weather Degradation} 
        & Sunny / Cloudy & $0$ \\
        & Rain / Snow & $+1$ \\
        & Sandstorm / Fog & $+3$ \\ 
        \midrule
        
        \multirow{3}{*}{C. Illumination} 
        & 07:00 -- 17:00 (Daylight) & $0$ \\
        & 06:00--07:00 \& 17:00--18:00 (Twilight) & $+1$ \\
        & 18:00 -- 06:00 (Night) & $+2$ \\ 
        \midrule
        
        D. Victim Count & $N$ victims ($N \ge 1$) & $+N$ \\ 
        \midrule
        
        \multirow{3}{*}{E. Strong Clue Assistance} 
        & Presence of Tent & $-1$ \\
        & Presence of Bonfire & $-2$ \\
        & Presence of Flare & $-3$ \\ 
        \bottomrule
    \end{tabularx}
    \label{tab:difficulty}
\end{table}

As illustrated in \Cref{tab:difficulty}, we establish a quantitative scoring metric to evaluate the difficulty of each embodied search and rescue task. Specifically, we utilize the 25th, 50th (median), and 75th percentiles of the average distance between the starting point and the victims to formulate the distance-based scoring criteria. Building upon this, we systematically incorporate environmental factors—including weather, illumination conditions, the number of victims, and the presence of special clues—assigning corresponding positive or negative difficulty scores. Ultimately, these individual components are aggregated to compute the total difficulty score for task $T_i$:
\begin{equation}
    S(T_i) = S_\text{dist} + S_\text{weather} + S_\text{light} + S_\text{count} + S_\text{clue}.
    \label{eq:difficulty}
\end{equation}
Based on the final score $S(T_i)$, we categorize the tasks within the benchmark into four distinct difficulty levels: Simple ($S \le 3$), Medium ($3 < S \le 5$), Hard ($5 < S \le 7$), and Extreme ($S > 7$).

\section{Calculation of the Success Rate}
\label{sec:sr}
To accurately compute the Success Rate (SR), the association between the UAV's $N$ predicted coordinates and the $M$ actual targets is formulated as a linear assignment problem. First, the spatial distance between each prediction-target pair is calculated to construct an $N \times M$ cost matrix. To resolve ambiguous associations, such as multiple predicted coordinates clustering around a single actual target, the Hungarian algorithm is applied to this cost matrix. This guarantees an optimal one-to-one bipartite matching that minimizes the total assignment distance. A prediction is considered to have successfully located a target if and only if:
\begin{itemize}
\item It is optimally matched to the target by the algorithm.
\item The distance between them is less than the predefined error threshold $E$.
\end{itemize}

\section{Details of the Four Reference Areas}
To replicate realistic search and rescue (SAR) scenarios, we selected four regions in China, each characterized by distinct topographical features and a high incidence of SAR events, as references for constructing our simulation environment. These regions represent four typical wilderness terrains: alpine meadows, desert and Gobi, snow-capped peaks, and coastal areas. Detailed descriptions of the four locations are provided below:

\vspace{1em}

\noindent
\begin{minipage}[c]{0.35\textwidth} 
    \centering
    \includegraphics[width=\linewidth]{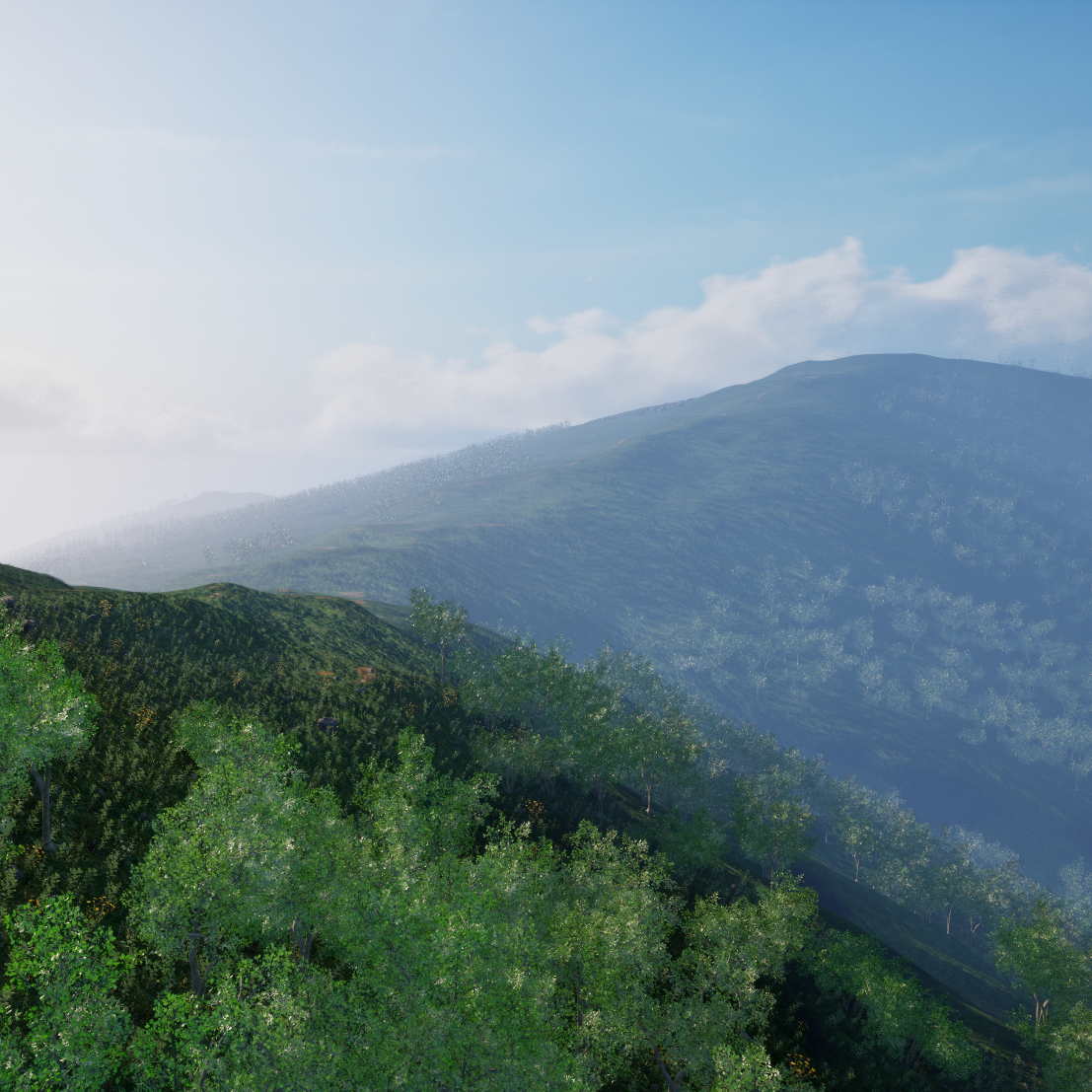}
\end{minipage}%
\hfill
\begin{minipage}[c]{0.60\textwidth} 
    \textbf{Map 1: Aotai Trail} \\
    Located along the main ridge of the Qinling Mountains, this is one of the most well-known hiking routes in China. The trail traverses uninhabited areas including forests, alpine meadows, blockfields, and mountain ridges. Frequent extreme weather and a highly variable climate result in a high accident rate. Since 2012, at least 58 individuals have been reported missing or deceased along this route. The simulation environment reconstructs a 2 km × 2 km area surrounding the "2800 Campsite" on the trail.
\end{minipage}

\vspace{1em}

\noindent
\begin{minipage}[c]{0.35\textwidth}
    \centering
    \includegraphics[width=\linewidth]{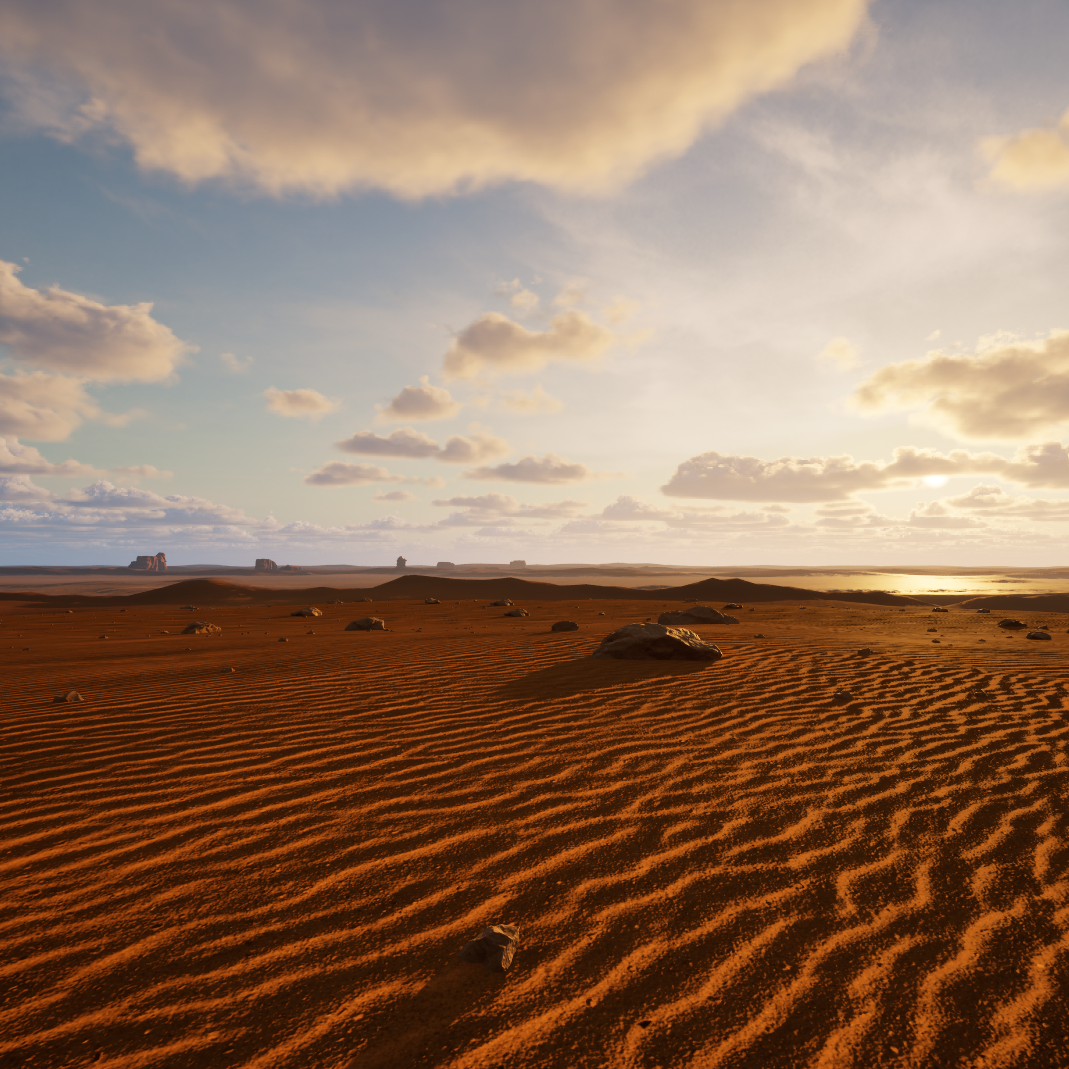}
\end{minipage}%
\hfill
\begin{minipage}[c]{0.60\textwidth}
    \textbf{Map 2: Lop Nur} \\
    Situated in the Tarim Basin, Lop Nur is a desiccated salt lake and one of China's most widely recognized uninhabited regions. The terrain is predominantly desert and Gobi. It remains arid year-round and is frequently subjected to sandstorms. Historically, numerous accidents involving scientific expeditions and explorations have occurred here. The simulation environment models a 5 km × 5 km area centered around the tomb of explorer Yu Chunshun.
\end{minipage}

\vspace{1em}

\noindent
\begin{minipage}[c]{0.35\textwidth}
    \centering
    \includegraphics[width=\linewidth]{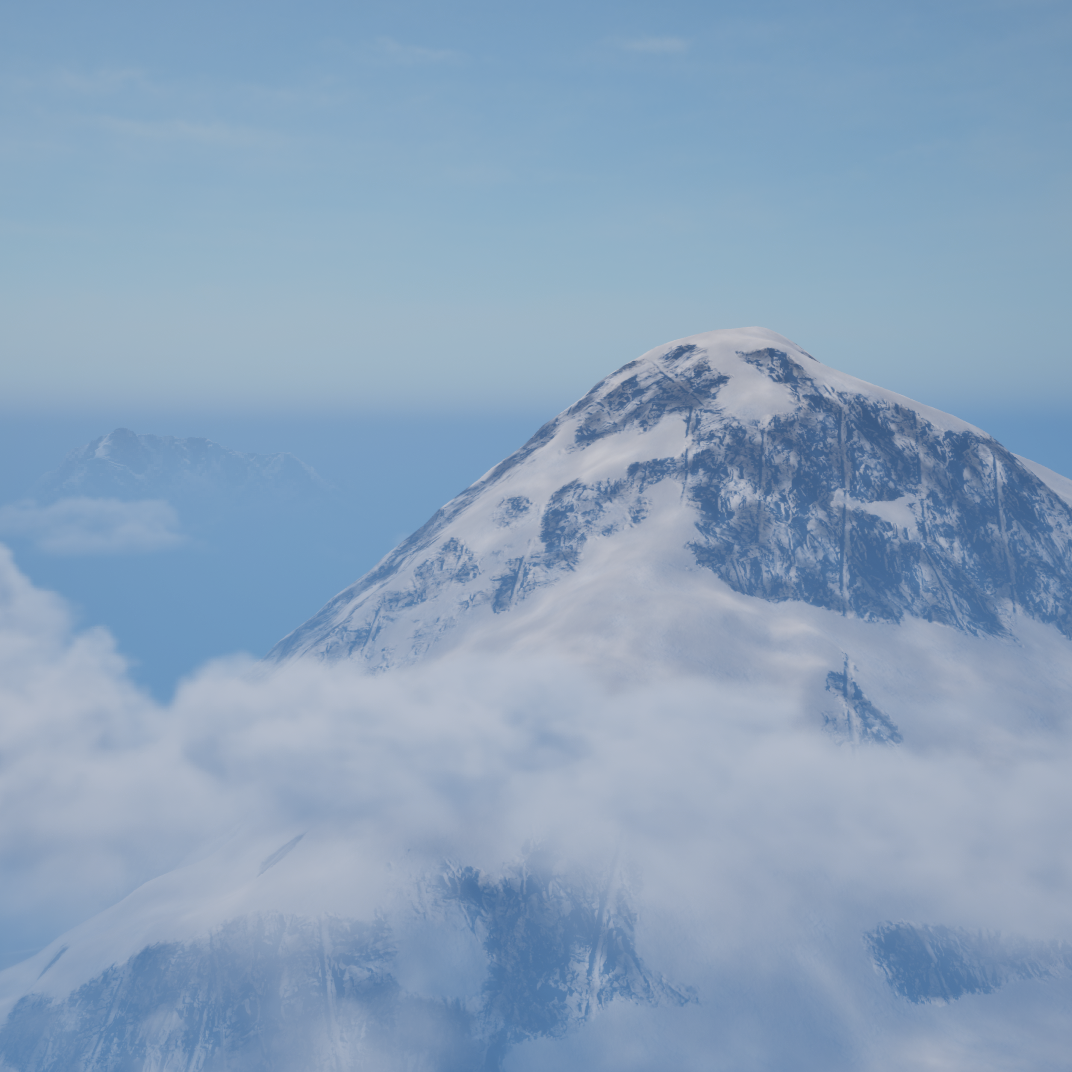}
\end{minipage}%
\hfill
\begin{minipage}[c]{0.60\textwidth}
    \textbf{Map 3: K2} \\
    As the second-highest peak in the world, K2 is widely considered one of the most difficult mountains to summit. The peak experiences extreme conditions, with minimum temperatures dropping to -50°C and wind speeds reaching 50 m/s, alongside a high frequency of avalanches. Since 1954, 92 fatalities have been recorded, resulting in a fatality rate of approximately 20\%. The simulation environment incorporates a 2 km × 2 km area surrounding the summit.
\end{minipage}

\vspace{1em}

\noindent
\begin{minipage}[c]{0.35\textwidth}
    \centering
    \includegraphics[width=\linewidth]{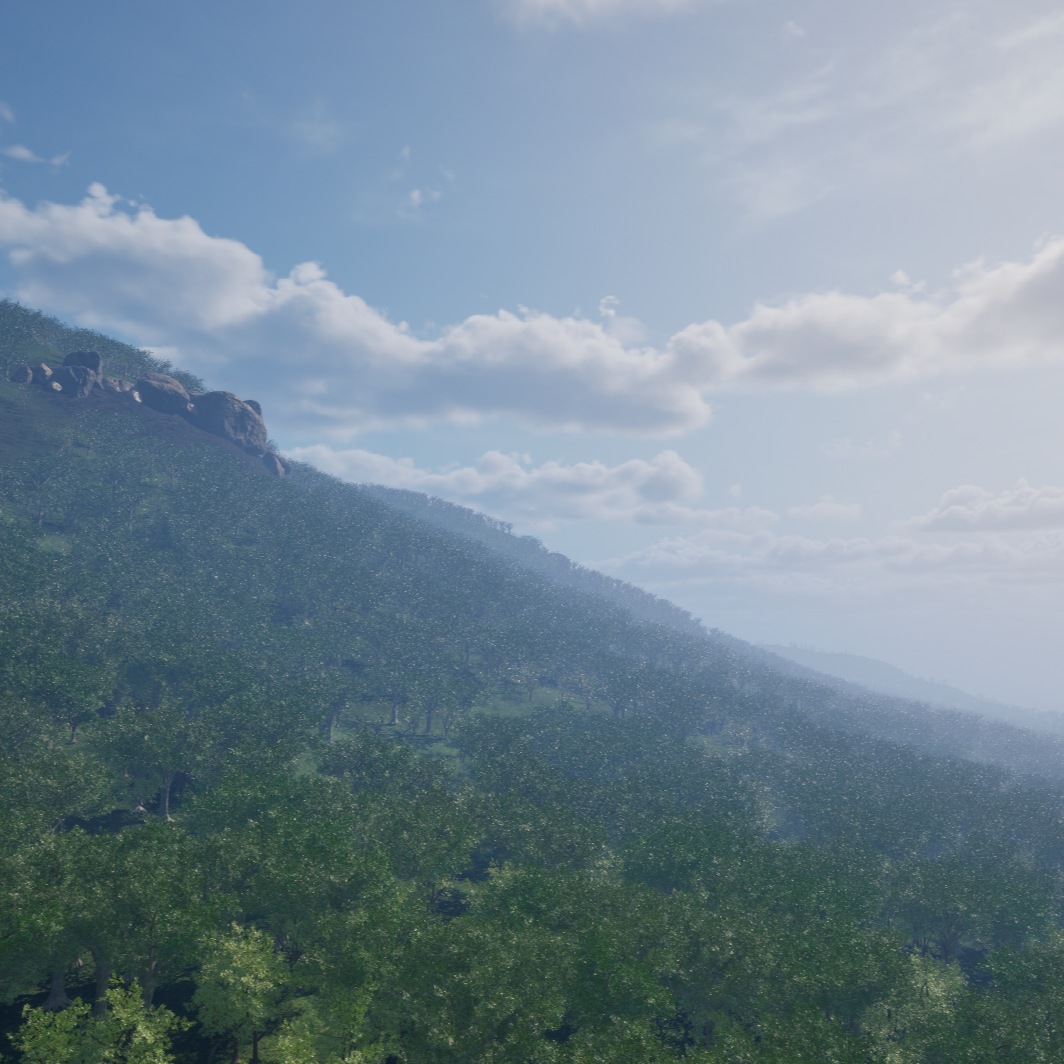}
\end{minipage}%
\hfill
\begin{minipage}[c]{0.60\textwidth}
    \textbf{Map 4: Dapeng Peninsula} \\
    Located near Shenzhen, China, this peninsula is a popular tourist destination characterized by hilly terrain, forests, and coastal landscapes. Due to its proximity to the urban center, many hikers attempt to explore its undeveloped areas without adequate preparation, leading to frequent accidents. Since 2025, at least five fatalities have occurred in this region. The simulation environment covers a 2 km × 2 km area around Wanglanggui.
\end{minipage}

\section{Calculation of the Clue Discovery Score}
\label{sec:cds}
To accurately compute the Clues Discovery Score (CDS), we evaluate the UAV agent's reported clues across two distinct tiers of accuracy: pure spatial localization and exact semantic matching. A ground-truth clue is categorized based on the following conditions:

\begin{itemize}
    \item \textbf{Spatial Localization ($C_\text{loc}$):} The clue is counted as spatially located if the Euclidean distance between the reported coordinates and the actual ground-truth coordinates is strictly less than a predefined distance threshold $E$. This acknowledges the practical utility of identifying suspicious regions in real-world SAR operations, regardless of semantic correctness.
    \item \textbf{Exact Semantic Matching ($C_\text{exact}$):} The clue achieves an exact match if it strictly satisfies the spatial localization condition (distance $< E$) \emph{and} a Large Language Model (LLM) identifies a valid semantic match between the reported text and the actual ground-truth clue name.
\end{itemize}

The prompt template utilized for this LLM-based semantic evaluation is illustrated in \Cref{tab:prompt}.

\begin{table}[h]
\centering
\caption{The prompt template used for LLM-based semantic matching evaluation.}
\label{tab:prompt}
\begin{tcolorbox}[colback=gray!5!white,colframe=gray!50!black,width=0.95\textwidth,arc=4mm,boxrule=1pt]
\small
\textbf{Model:} qwen3-max \\
\textbf{Prompt:} \\
Role: You are an expert judge for a Search and Rescue (SAR) mission. \\
Task: Determine if the objects reported by a drone match the actual ground truth clues. \\

\textbf{Input Data:}
\begin{itemize}
    \setlength{\itemsep}{0pt}
    \setlength{\parskip}{0pt}
    \item Ground Truth List: \{ground\_truth\_cues\}
    \item Drone Reported List: \{agent\_cues\}
\end{itemize}

\textbf{Matching Logic:}
\begin{enumerate}
    \setlength{\itemsep}{0pt}
    \setlength{\parskip}{0pt}
    \item Perform semantic matching. (e.g., "red bag" matches "Backpack", "fire" matches "Campfire").
    \item A drone report matches a ground truth if they refer to the same physical object.
    \item Multiple reports might refer to the same ground truth clue; ensure you count unique ground truth matches.
\end{enumerate}

\textbf{Output Format (Strict JSON only, no conversational filler):} \\
\{ \\
\indent "matches": [ \\
\indent \indent \{"agent\_cue":"reported\_term","gt\_cue":"matched\_ground\_truth\_term"\} \\
\indent ], \\
\indent "matched\_gt\_count": <number-of-unique-ground-truth-clues-found> \\
\}
\end{tcolorbox}
\end{table}

\section{Justification of the Static Snapshot Formulation}
\label{sec:snapshot}
While real-world SAR missions involve dynamic victims, modeling the environment as a series of static temporal snapshots is a highly justifiable approximation for Embodied AI evaluation, grounded in both kinematics and SAR operational priors:

1) \textbf{Behavioral Prior:} In wilderness SAR, high-priority targets are frequently incapacitated or adopt a "stay-in-place" survival strategy, making their overall displacement negligible during a single search sortie.

2) \textbf{Kinematic Dominance ($v_d \gg v_h$):} The relative search velocity $v_{rel} \approx v_d - v_h \cos\theta$ is overwhelmingly dominated by the UAV's velocity $v_d$. The target's movement vector acts as an marginal factor to the search geometry.

3) \textbf{The Improbability of "Slip-Through" Evasion:} The only scenario where a static approximation fails is the "perfect miss"—where a victim crosses from an unsearched region into an already-searched region without intersecting the UAV's sensor footprint. Geometrically, for this adversarial boundary crossing to occur, the velocity ratio must satisfy $v_h/v_d \ge \lambda R_s / L_B$ (where $R_s$ is the sensor radius and $L_B$ is the characteristic search length). Given typical UAV speeds ($>5$ m/s) and rough-terrain human walking speeds ($<0.5$ m/s), this condition is strictly violated in standard SAR operations. Thus, the probability of missing a target purely due to its active movement is mathematically negligible under random walk assumptions.

\begin{figure*}[h]
    \centering
    \includegraphics[width=1.0\linewidth]{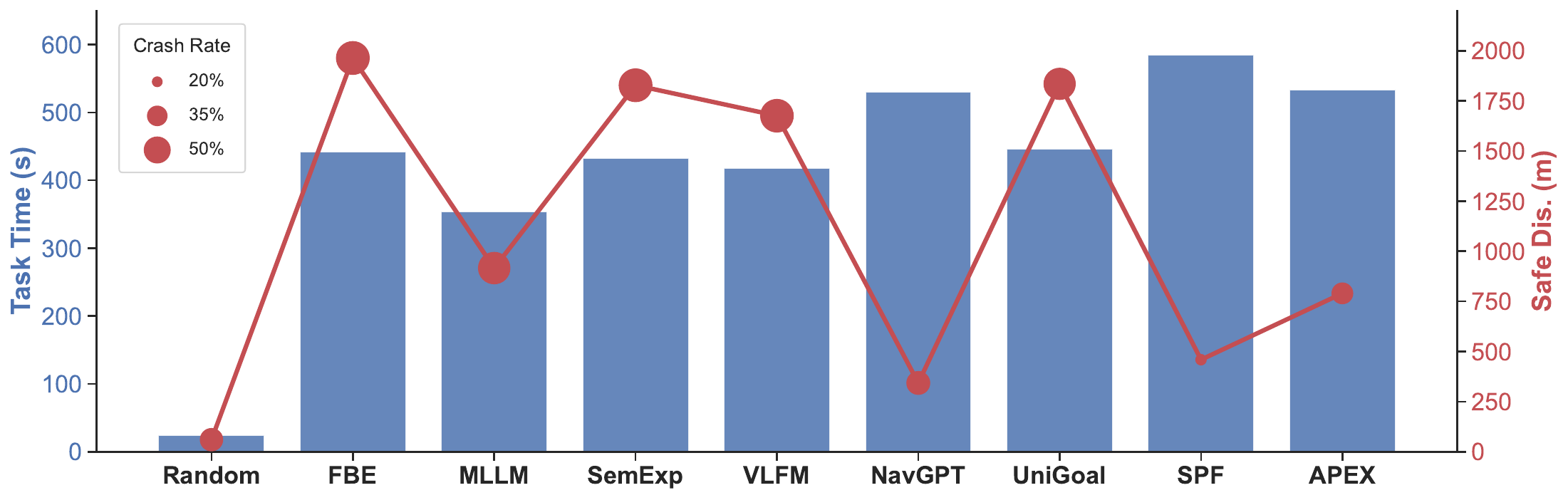}
    \caption{\textbf{More experimental results analysis.} Crash rate, task time, and safe flight distance of different baseline methods. Larger red dots indicate higher crash rates. The results reveal a clear trade-off between search duration and flight safety: methods with stronger exploration ability often require longer task time. Meanwhile, the crash rates across most baselines indicate that safe long-horizon UAV operation remains a major challenge in ESARBench.}
    \label{fig:plot}
    \vspace{-10pt}
\end{figure*}

\section{More Results and Discussion}
\label{sec:discussion}
\begin{figure*}[h]
    \centering
    \includegraphics[width=1.0\linewidth]{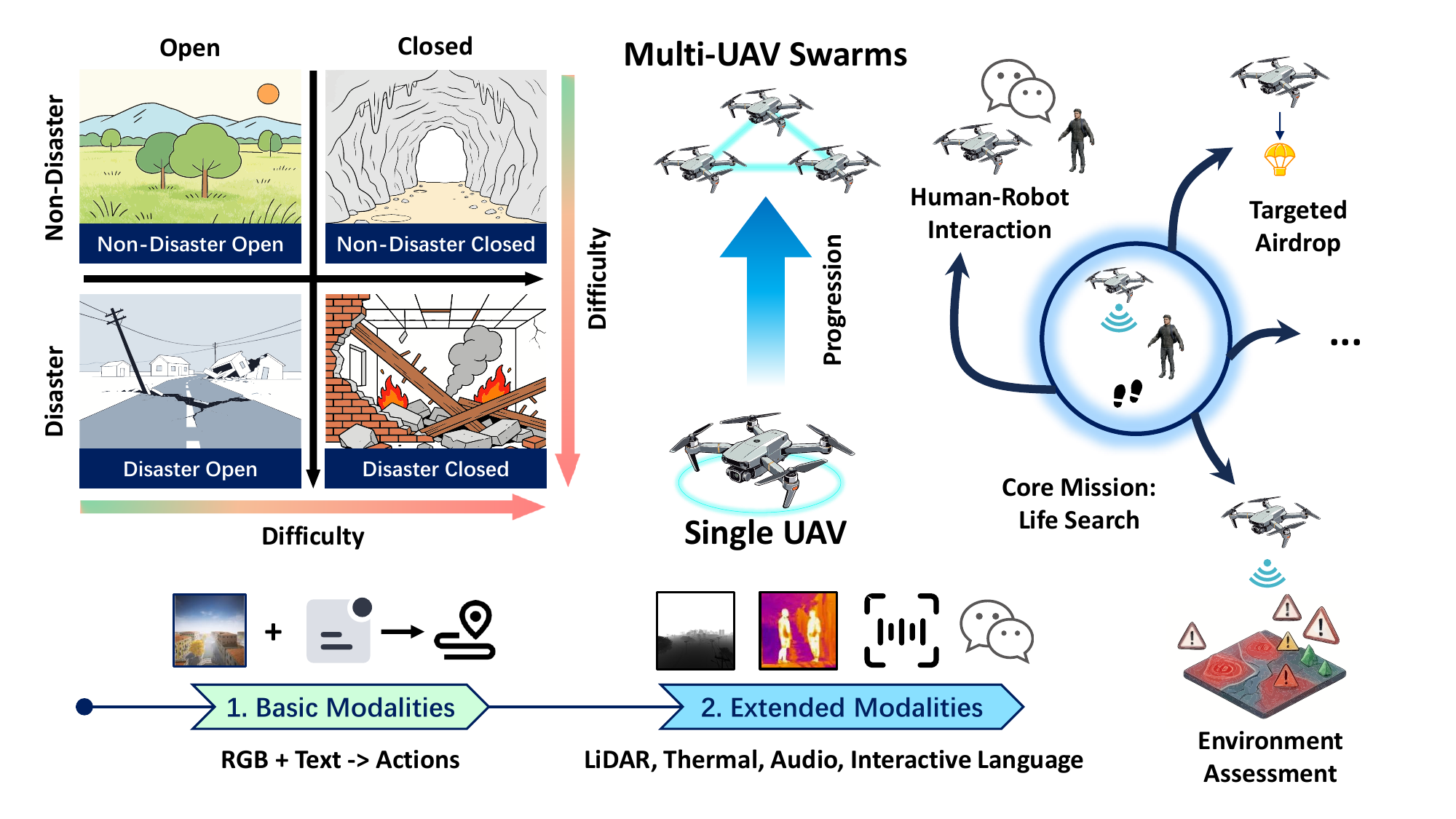}
    \caption{\textbf{Future Development of Embodied Search and Rescue (ESAR).} This diagram illustrates the field's future development across four dimensions: (1) Operational Scenarios: Scaling from stable, unconstrained open spaces to unpredictable, restricted disaster zones. (2) Architectural Progression: Advancing from robust single-UAV to multi-UAV swarm collaboration. (3) Task Formulation: Diverging from the core mission of active search into several auxiliary tasks. (4) Multi-modal Fusion: Evolving from basic sensory inputs (RGB and discrete text) to extended modality combinations.}
    \label{fig:overview}
\end{figure*}

While this work establishes the inaugural benchmark for ESAR, we conceptualize a broader roadmap that scales in complexity across scenarios, tasks, and architectures. As illustrated in \Cref{fig:overview}, future deployment environments are envisioned to progress from unconstrained open spaces to restricted enclosed areas, and from stable non-disaster events to dynamic disaster zones. Similarly, agent capabilities must evolve from foundational visual trace searching to auxiliary objectives like risk assessment and targeted airdrops, while architectures advance from robust single-UAV autonomy to collaborative multi-UAV swarms.

In ESARBench, we concentrated on the initial category: life searching in open, non-disaster environments. This scoping allowed us to isolate the core challenges of multimodal perception, long-horizon reasoning, and spatial exploration. However, bridging the gap to real-world deployment  requires extending these axes, including the integration of  thermal/audio modalities for hidden victim detection and introducing environmental volatility to test robustness.

The experiment ran for about 140 hours on a single A100 GPU, costing about 8G VRAM.

\section{Real-World Rescue Cases}
\label{sec:realcase}
\begin{figure*}[h]
    \vspace{-10pt}
    \centering
    \includegraphics[width=1.0\linewidth]{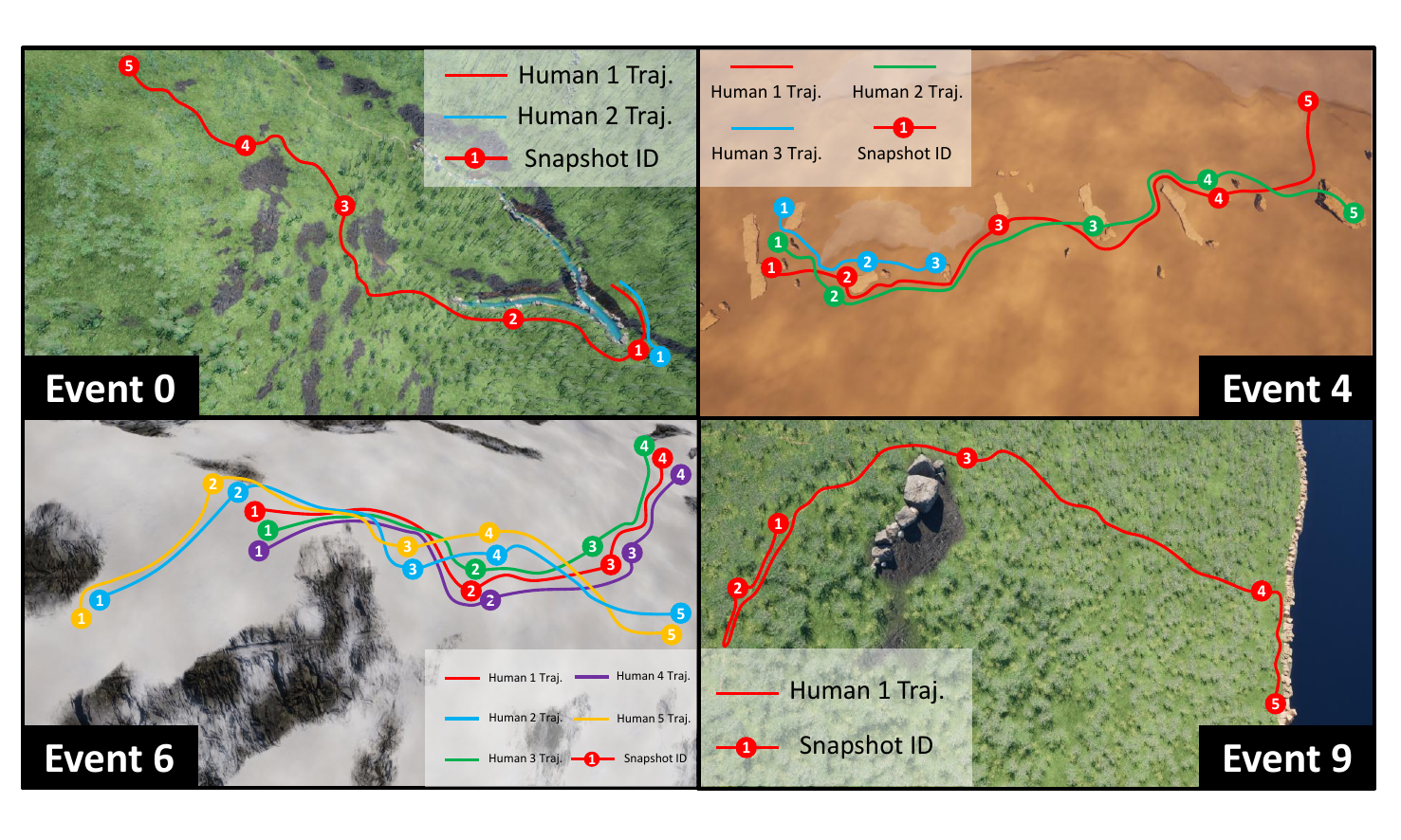}
    \vspace{-20pt}
    \caption{\textbf{Visualization of four representative event examples.} An event represents a complete, longitudinal real-world search and rescue incident that unfolds over an extended period. Each event is discretized  into multiple static time snapshots.}
    \label{fig:realcase}
    \vspace{-5pt}
\end{figure*}
\Cref{fig:realcase} illustrates some of the real-world search and rescue (SAR) incidents utilized as references for task generation. Detailed descriptions of these events are provided below:

1) On September 25, 2021, a hiker Wu began a travel along the Aotai Trail. Three days later, he got injury while descending from the 2800 Campsite and agreed with his teammates to remain in place to wait for rescue. However, several days later, Wu unexpectedly ascended toward the summit, resulting in a missed encounter with the rescue team. Although he was discovered by another hiker on October 3, he died of hypothermia while the second hiker left to seek assistance.

2) On June 11, 1996, the renowned explorer Yu began a solo traverse of Lop Nur. During the expedition, he encountered a severe sandstorm that forced him off his intended route, preventing him from accessing the supply caches pre-positioned along his planned path. Several days later, Yu perished near a trail intersection; the cause of death was  dehydration and heat exhaustion.

3) On June 20, 1986, five mountaineers successfully summited K2. However, during their descent, the team encountered a severe blizzard. While three climbers rapidly descended to a secure area, the remaining two became disoriented in the storm and ultimately died from a fall.

4) On October 16, 2025, a hiker Zhong proceeded to explore an undeveloped area of the Dapeng Peninsula without carrying any professional equipment. Telemetry data from his sports watch indicated a suspected wildlife attack during the hike. He subsequently descended to the coastal area, where he was ultimately found deceased. The exact cause of death remains undetermined.



\end{document}